%% file: arxiv.tex
\PassOptionsToPackage{table,dvipsnames}{xcolor}
\PassOptionsToPackage{numbers}{natbib}
\documentclass[10pt, a4paper, logo]{googledeepmind} %
\usepackage{rotating}
\usepackage{algpseudocode}
\setlist[itemize]{left=0pt}

\usepackage{fancybox}

\input{math_commands.tex}

\usepackage{multirow}   % For multirow cells
\usepackage{tabularx}   % For tables with adjustable width
\usepackage{makecell}   % For line breaks inside cells

\usepackage{etoc}
\usepackage{xspace}
\hypersetup{pdftitle={Reward Budgeting Reduces Premature Convergence in Reinforcement Learning for LLM Reasoning},
  pdfauthor={Mengni Jia, Mengyu Zhou, Xiaoxi Jiang, Guanjun Jiang},
  linkcolor=black,citecolor=blue!50!black,urlcolor=blue!60!black}
\usepackage{float}

\usepackage{longtable} % allows multipage tables
\usepackage{array}     % for \arraystretch
\usepackage{circledsteps}

\usepackage{natbib}
\usepackage{subcaption}
\usepackage{fvextra}
\usepackage{mathtools}
\usepackage{nicefrac}       % compact symbols for 1/2, etc.
\definecolor{gainGreen}{HTML}{267A46}
\newcommand{\cmark}{\textcolor{green!60!black}{\ding{51}}}
\newcommand{\xmark}{\textcolor{red!70!black}{\ding{55}}}
\usepackage{tikz}
\usetikzlibrary{arrows.meta, positioning}
\usepackage{wrapfig}
\usepackage{epigraph}
\usepackage{algorithm}
\definecolor{oursgray}{gray}{0.95}

\theoremstyle{plain}
\newtheorem{theorem}{Theorem}[section]

\newtheorem{lemma}[theorem]{Lemma}
\newtheorem{corollary}[theorem]{Corollary}
\theoremstyle{definition}

\newtheorem{assumption}[theorem]{Assumption}
\theoremstyle{remark}

\usepackage[capitalize,noabbrev]{cleveref}

\title{Reward Budgeting Reduces Premature Convergence in Reinforcement Learning for LLM Reasoning}

\author[2]{Mengni Jia}
\author[1]{Mengyu Zhou\textsuperscript{\dag}}
\author[1]{Xiaoxi Jiang}
\author[1]{Guanjun Jiang}
\affil[1]{Qwen Business Unit, Alibaba}
\affil[2]{University of Cambridge\newline\textsuperscript{\dag}Corresponding author: \href{mailto:zhoumengyu.zmy@alibaba-inc.com}{zhoumengyu.zmy@alibaba-inc.com}}

\begin{abstract}
RL for LLM reasoning often plateaus early as its policy sharpens. Policy entropy is studied as both a diagnostic and a control signal, but increasing it alone does not reliably change the eventual plateau in our experiments. We analyze RL's implicit target policy using a sharpness factor $\kappa$ and effective counterpart $\kappa_{\mathrm{eff}}$ that describe how reward or advantage gaps translate into probability-ratio gaps. We introduce \emph{reward budgeting}, a plug-in advantage construction assigning a fixed total reward budget to each prompt. For binary rewards and independent on-policy rollouts, its expected unclipped score-function update is the exact gradient of a finite-$G$ concave objective with diminishing returns; its advantage gap decreases as more rollouts become correct. For general rewards, a shifted log-mean interpretation holds asymptotically under the stated regularity conditions. On Qwen3-8B and Llama3.2-3B-Instruct, reward budgeting consistently improves RL baselines, with gains up to +12.50 on AMC23 and +10.00 on AIME24 and AIME25. These results support an update-dynamics view of premature convergence and substantial gains without explicitly preserving entropy. Code is available at \url{https://github.com/Qwen-Applications/RewardBudget}.
\end{abstract}

\begin{document}
\maketitle

\section{Introduction}
\label{sec:intro}

Policies trained with RL for LLM reasoning often improve rapidly at first and then enter a plateau where further training brings little gain. Such a plateau need not reflect an intrinsic limit of the model or data: optimization choices can change the performance reached under otherwise similar training conditions. For example, \cite{khatri2025art} report that, in their scaling study, GSPO~\citep{zheng2025groupsequencepolicyoptimization} and CISPO~\citep{minimax2025minimaxm1scalingtesttimecompute} reach higher fitted asymptotic performance than DAPO~\citep{yu2025dapoopensourcellmreinforcement}. This motivates us to distinguish \emph{natural convergence} from \textbf{algorithm-induced premature convergence}: training may stop making useful progress, even when a different algorithm can reach a better solution.

\begin{wrapfigure}{r}{0.44\linewidth}
    \centering
    \includegraphics[width=\linewidth]{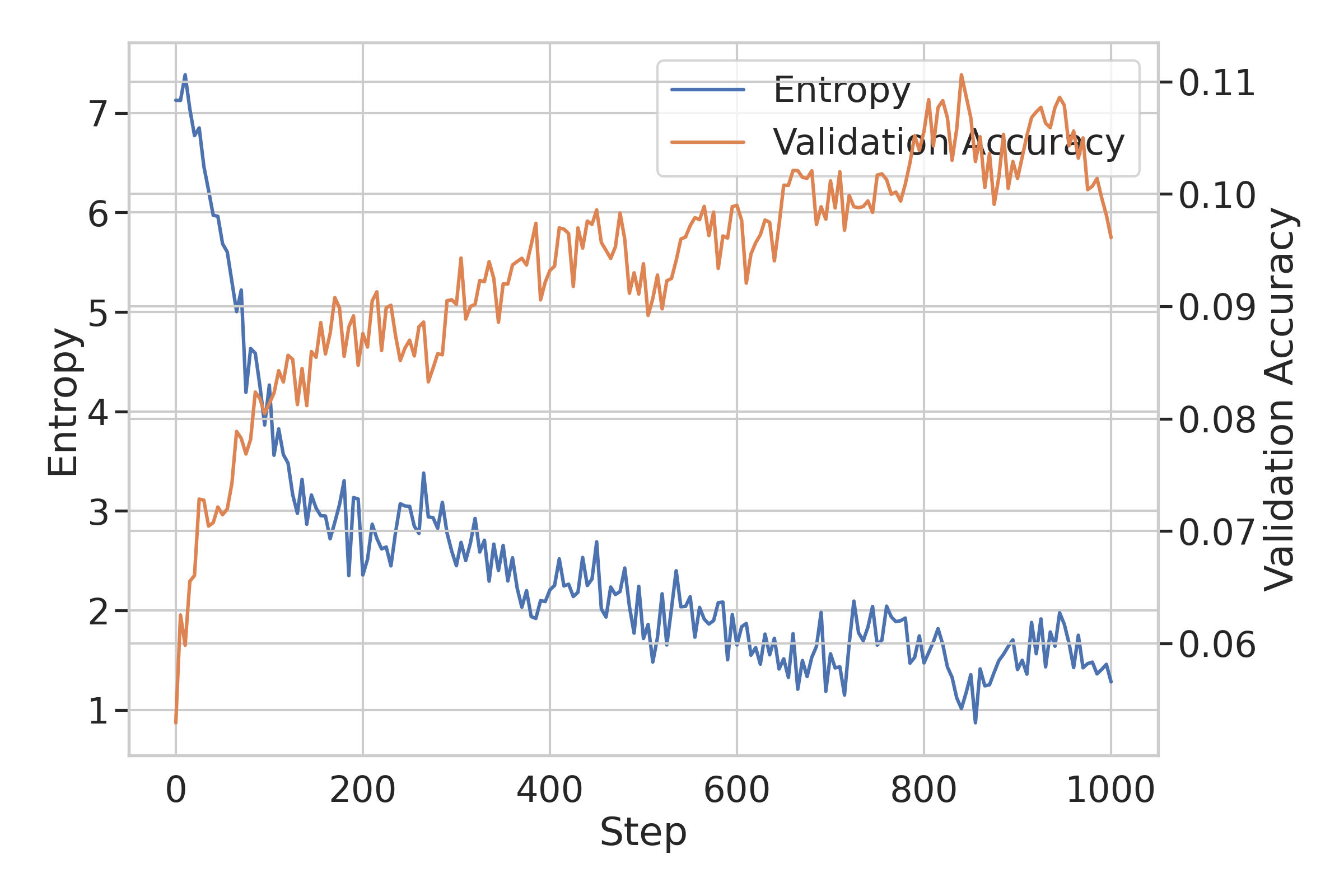}
    \caption{Average training entropy and validation accuracy across 12 Llama3.2-3B-Instruct runs.}
    \label{fig:entropy_collapse}
\end{wrapfigure}

This phenomenon has been commonly referred to as ``entropy collapse''. This name arises from the \textbf{observation} that the policy seems to stop learning and converge prematurely once its entropy drops to a relatively low level \citep{cui2025entropymechanismreinforcementlearning}; see Figure~\ref{fig:entropy_collapse}. Consequently, many proposed remedies focus on \emph{explicit} entropy control, typically by encouraging higher entropy or keeping it within a prescribed range~\citep{cui2025entropymechanismreinforcementlearning, NEURIPS2025_a797c2d2, cheng2025reasoningexplorationentropyperspective, pmlr-v306-tan26e, jiang2025rethinkingentropyregularizationlarge, hao2025onpolicyrloptimalreward,he2025skyworkopenreasoner1, zhang-etal-2026-revisiting, wang2025arbitrary, park2025cliplowincreasesentropycliphigh, xu2026epoentropyregularizedpolicyoptimization,petrenko2026entropypreservingreinforcementlearning, hao-etal-2026-rethinking, lei2025revisitingllmreasoninginformation, zhang2025freelunchrethinkinginternal, yu2025dapoopensourcellmreinforcement, su-etal-2026-ce, yang2026entropicstablelongtermtraining, zhang-etal-2026-revisiting, li2025curecriticaltokenguidedreconcatenationentropycollapse, jin-etal-2026-revisiting}; see~\ref{app:related_work} for a discussion. 
These studies provide substantial evidence that entropy is related to RL training dynamics. Here, we ask a narrower question: \emph{is changing entropy itself sufficient to change the eventual performance plateau?}

Our results suggest caution. First, the familiar entropy--accuracy pattern is often observed after averaging across runs, while individual methods have different entropy \mbox{trajectories}~(\S\ref{sec:the_smoothing_trap}). Second, in the intervention study of \S\ref{sec:entropy_is_not_the_cause}, two different entropy-promoting interventions substantially change entropy during early training, yet the trajectories return toward the same baseline behavior after the interventions are removed and do not reach a better final plateau. This does not imply that entropy-based methods cannot be effective: such methods may alter optimization in useful ways beyond simply raising entropy. It does suggest that, in our setting, entropy promotion alone is not a sufficient intervention for premature convergence.

Conceptually, entropy measures how dispersed the policy distribution is, but it does not indicate whether the resulting responses are correct or whether their diversity is useful. A high-entropy policy can produce diverse but unproductive outputs, while a low-entropy policy can be confidently correct. We therefore view entropy as an informative diagnostic of policy concentration rather than, by itself, an explanation of why training plateaus. For this reason, we use the term \emph{premature convergence} rather than \emph{entropy collapse} throughout the paper.

Once we question whether entropy is the primary cause behind the plateau, a natural question follows: if we do not control entropy directly, how can we sustain meaningful learning signals so that training can continue to realize incremental gains? In fact, reduced entropy (and the resulting sharper policy distribution) is expected, since it indicates that the policy is learning and becoming more confident on the training data. We just wish to avoid improper sharpening that locks the policy into cheap pseudo-modes and leaves little room for further gains. In generative modeling, mode-seeking behavior can depend on the divergence being optimized~\citep{pmlr-v206-ting-li23a}. This motivates examining objective-dependent sharpening in our RL setting. However, simply delaying convergence does \emph{not} guarantee stronger performance: gentler dynamics can slow training and require more data for sufficient learning. What we need is a mechanism that maintains gentler training dynamics to prevent premature sharpening, while adaptively relaxing control when genuine learning signals appear.

% To understand the underlying mechanism, we delve into RL's sharpening nature in \S\ref{sec:sharpness_analysis}. We characterize the training dynamics through the optimal-policy perspective, where RL can be viewed as moving the current policy toward a reward-weighted target policy $\pi^*$. This analysis isolates a \emph{sharpness factor} $\kappa$, which directly controls how advantage gaps are amplified into probability-ratio gaps between positive and negative responses. 
% Subsequent diagnostics, including controlled studies in~\S\ref{sec:sharpness_factor_controlled_studies}, dis-aggregation studies in~\S\ref{sec:the_smoothing_trap}, and deconfounding regression in~\S\ref{app:deconfounding_regression}, then compare $\kappa$ to entropy, and suggest that $\kappa$ is a more mechanistically grounded quantity than entropy~(\S\ref{sec:sources}). 

To describe the training dynamics more directly, we analyze RL through its implicit target policy in \S\ref{sec:sharpness_analysis}. For KL-regularized RL, the optimal target takes the reward-weighted form
$\pi^*(\gamma)\propto\pi_{\mathrm{ref}}(\gamma)\exp(R(\gamma)/\beta)$.
This view isolates a \emph{sharpness factor} $\kappa$, which determines how strongly reward gaps are amplified into probability-ratio gaps between higher- and lower-reward responses. We further define an effective version, $\kappa_{\mathrm{eff}}$, for practical algorithms with normalized advantages, clipping, and other implicit constraints. In this sense, $\kappa$ and $\kappa_{\mathrm{eff}}$ describe the sharpening pressure induced by the update itself, rather than the resulting uncertainty of the policy.

Motivated by this, we introduce \textbf{reward budgeting}. For each prompt, we allocate a fixed total reward budget across sampled rollouts according to their positive reward gaps. In the binary-reward setting, if $k$ out of $G$ rollouts are correct, each correct rollout receives a reward share proportional to $1/k$. As a result, the advantage gap decreases as a prompt becomes easier, producing stronger sharpening when successes are rare and gentler sharpening when successes are already common. Importantly, this behavior emerges without explicitly constraining policy entropy. We also discuss reference-free optimization in \S\ref{sec:prior_never_fades}: removing a fixed KL reference avoids permanently anchoring the target policy to a static prior, while importance-ratio clipping can still provide local regularization.

Theoretically, reward budgeting admits a principled interpretation rather than a heuristic tweak. First, it regulates the $\kappa$ trajectory, distinguishing it from most existing methods. As more rollouts become correct, each correct rollout receives a smaller share of the reward, which implicitly reduces the sharpness factor $\kappa$ and automatically relaxes the sharpening pressure. Second, under binary rewards, which match our main RLVR experiments, the expected update of the proposed estimator is exactly the gradient of a finite-$G$ concave objective with diminishing returns~(\S\ref{app:oo_binary}). Thus, repeatedly solving an already-easy prompt contributes progressively less marginal benefit than under a linear expected-reward objective. 
Under general rewards, the gap-normalized estimator admits a shifted log-mean interpretation in the large-$G$ limit under suitable support assumptions~(\S\ref{app:oo_general}).
% Importantly, neither of these properties are imposed by hand; they both emerge naturally from the training dynamics of reward-budgeted algorithms. This perspective makes reward budgeting conceptually clean and explains how reward budgeting mechanistically mitigates premature convergence.

Empirically, our method substantially improves on Qwen3-8B and Llama3.2-3B-Instruct across various RL baselines (up to +12.50 on AMC 23 and +10.00 on AIME24 \& AIME25).
The gains do not require higher entropy; after applying reward budgeting, entropy often decreases while final performance improves.
Collectively, these results suggest that regulating algorithm-induced sharpening is a practical way to mitigate premature convergence, while entropy remains a useful diagnostic rather than a necessary quantity to preserve.

\section{Related Work}
\label{app:related_work}

Premature convergence in RL for LLM reasoning is often accompanied by a rapid decline in policy entropy. Recent studies have examined this behavior from different perspectives, including entropy dynamics, token selection, reward shaping, and regularization. While these approaches share an interest in improving exploration, they differ in how entropy enters the policy update.

\paragraph{Entropy Dynamics and Token-Level Interventions.}
\cite{cui2025entropymechanismreinforcementlearning} establish an empirical relationship between policy entropy and validation performance, and analyze entropy changes through the covariance between action log-probabilities and policy updates. Their analysis motivates Clip-Cov and KL-Cov, which selectively constrain high-covariance token updates to slow entropy reduction. From another perspective, \cite{NEURIPS2025_a797c2d2} find that a small fraction of high-entropy tokens act as important forks in reasoning. They show that restricting policy-gradient updates to these tokens can match or outperform full-token training. Both works highlight the importance of where and how policy updates are applied. However, they focus on different signals: covariance characterizes contributions to entropy change, while token entropy identifies uncertain decision points. Our analysis instead examines how the RL objective and advantage construction determine the sharpening pressure of policy updates.

\paragraph{Entropy-Guided Credit Assignment.}
Other studies use entropy to improve the learning signal rather than directly regularizing the policy distribution. \cite{cheng2025reasoningexplorationentropyperspective} associate high-entropy regions with exploratory reasoning behaviors, such as reflection and correction. They add a clipped, gradient-detached entropy term to token advantages, encouraging such behaviors without directly optimizing an entropy bonus. Similarly, \cite{pmlr-v306-tan26e} propose GTPO and GRPO-S to address coarse-grained credit assignment in GRPO. Their methods use entropy to assign greater rewards to uncertain decisions in successful responses and stronger penalties to confident errors, at the token and sequence levels, respectively. These approaches show that entropy can help identify where learning signals may be more useful. Our method also modifies reward allocation, but uses the relative rewards of rollouts within each prompt rather than their entropy. The goal is to adjust how strongly each prompt contributes to learning as its sampled success rate changes.

\paragraph{Selective Entropy Regularization.}
Direct entropy regularization provides another way to encourage exploration, but increasing entropy indiscriminately can also destabilize training. \cite{jiang2025rethinkingentropyregularizationlarge} study this issue and propose SIREN, which combines top-$p$ action masking, high-entropy token selection, and a self-anchored regularizer. Rather than simply maximizing entropy, SIREN maintains a more controlled level of uncertainty over selected actions and positions. This distinction is important: the effectiveness of entropy-based methods can depend on which parts of the policy are regularized and how the updates are constrained, not only on the resulting entropy level.

\paragraph{Group-Relative Optimization and Reward Allocation.}
Our work is also related to the design of group-based RL objectives. GRPO~\citep{shao2024deepseekmathpushinglimitsmathematical} estimates advantages from relative rewards within each prompt group. Dr. GRPO~\citep{liu2025understanding} shows that group standard-deviation normalization can introduce difficulty-dependent weighting, and removes this normalization along with response-length normalization. DAPO~\citep{yu2025dapoopensourcellmreinforcement} improves training through several changes, including dynamic sampling that excludes groups with uniformly correct or incorrect rollouts. These methods improve advantage estimation or group selection. In contrast, reward budgeting changes how rewards are allocated within each group. Instead of assigning each successful rollout an independent unit reward, it distributes a fixed budget across the group according to reward gaps. Under binary rewards, this makes the advantage gap decrease as the number of correct rollouts increases.

\paragraph{Relation to Our Work.}
The above studies show that entropy can be useful for understanding and improving RL, but its role depends on how it is used. We examine a related question: whether increasing entropy itself is sufficient to change the final performance plateau. Our intervention experiments suggest that this is not reliable in the settings we study. We therefore focus on the sharpening pressure induced by policy updates, characterized by $\kappa$ and its effective counterpart $\kappa_{\mathrm{eff}}$. Reward budgeting provides a simple way to adjust this pressure through advantage construction, without explicitly controlling entropy. Moreover, under binary rewards, its expected policy-gradient update corresponds to a finite-$G$ concave objective with diminishing returns. Our results suggest that prompt-level reward allocation is another useful way to address premature convergence, alongside existing entropy-guided approaches.

\section{Sharpness Analysis}
\label{sec:sharpness_analysis}

Premature convergence is common in RL training, but its underlying dynamics remain unclear. In this section, we examine policy sharpening as a more direct description of RL updates than entropy alone. In \S\ref{sec:entropy_is_not_the_cause}, we show that increasing entropy alone does not reliably improve the final performance plateau in our settings. In \S\ref{sec:sources}, we derive a sharpness factor $\kappa$ from the RL objective, which characterizes how reward gaps are exponentially amplified in the implicit target policy. 
Finally, controlled experiments in \S\ref{sec:sharpness_factor_controlled_studies} and individual-run analyses in \S\ref{sec:the_smoothing_trap} show how sharpening strength relates to convergence dynamics, while averaging can obscure inconsistencies in the entropy--accuracy relationship.

% Finally, in \S\ref{sec:sharpness_factor_controlled_studies} and \S\ref{sec:the_smoothing_trap}, through controlled and dis-aggregated analyses, we show that $\kappa$ better explains convergence dynamics and that entropy collapse is partly amplified by averaging.

\subsection{Entropy}
\label{sec:entropy_is_not_the_cause}

Due to the observation that policies often plateau as their entropy drops, many remedies propose to control entropy. However, this correlation does not establish that entropy reduction causes the plateau. Before taking any action, we must first ask: is entropy truly the cause, or more like a correlate? This question is highly important, because treating the correlational signal as causal can lead to \emph{ineffective} interventions. Entropy may instead serve as a useful diagnostic of convergence dynamics, without being a sufficient target for intervention. 
We test this hypothesis as follows.

Let $H(\pi_{\theta})$ denote entropy on training data. We train Qwen3-8B in three runs that differ only by an \emph{early entropy-based intervention}. 
For the first $500$ steps, we use (1) \textbf{Baseline}: GRPO~\citep{shao2024deepseekmathpushinglimitsmathematical}. (2) \textbf{Add Entropy to Loss}: \(J_{\mathrm{RL}}(\theta) + 0.01\, H(\pi_{\theta})\), and (3) \textbf{Add Entropy to Reward}: $R(o)+0.01H(\pi_{\theta}(o))$. After $500$ steps, \emph{all} runs revert to the baseline and train for another 500 steps.

If early entropy promotion is sufficient to overcome premature convergence, the interventions should lead to higher post-intervention plateaus. Otherwise, the runs may return to similar performance once the interventions are removed.

\begin{figure}[htbp]
    \centering

    \begin{subfigure}{0.48\linewidth}
        \centering
        \includegraphics[width=\linewidth]{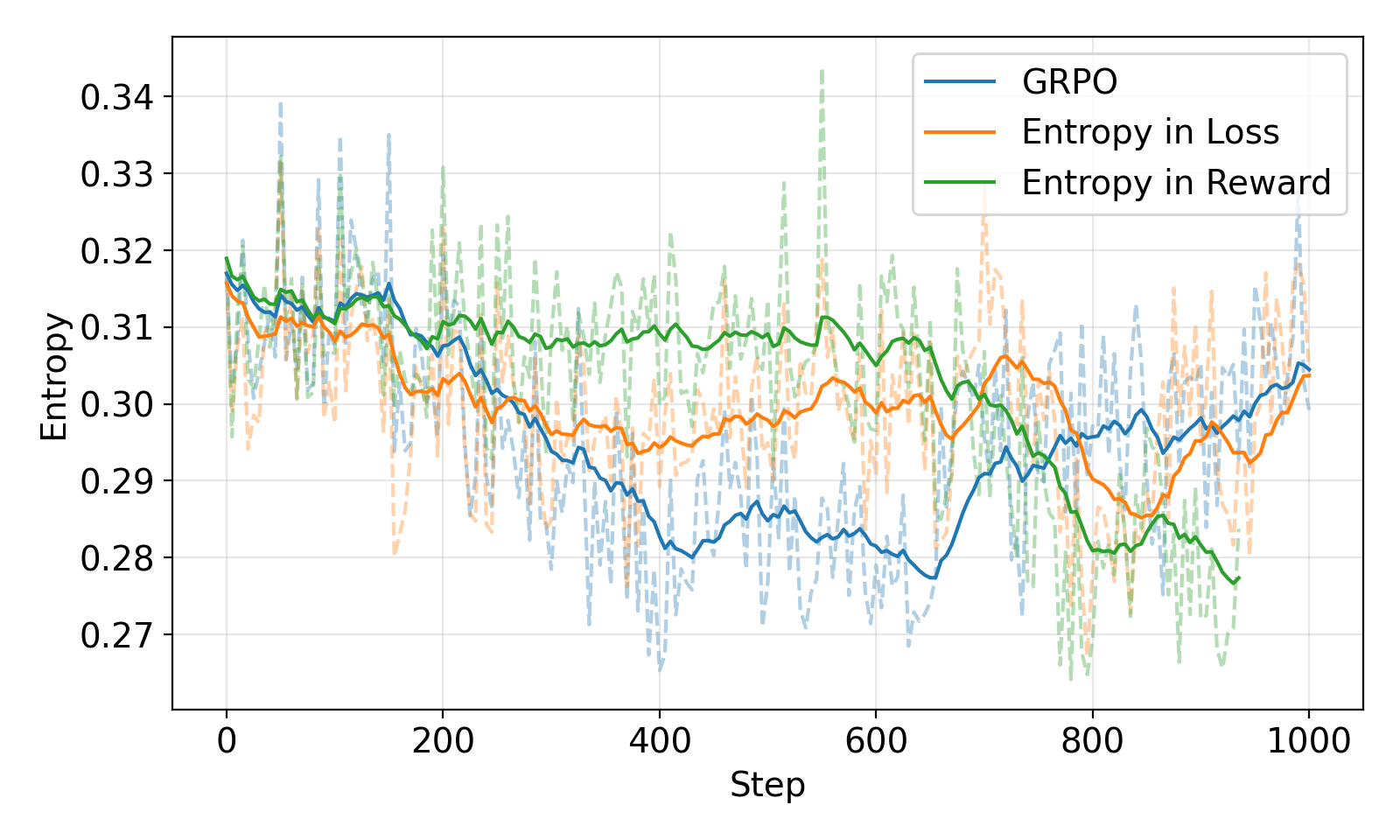}
        \caption{Entropy over training steps.}
        \label{fig:multiple_intervention_entropy}
    \end{subfigure}
    \hfill
    \begin{subfigure}{0.48\linewidth}
        \centering
        \includegraphics[width=\linewidth]{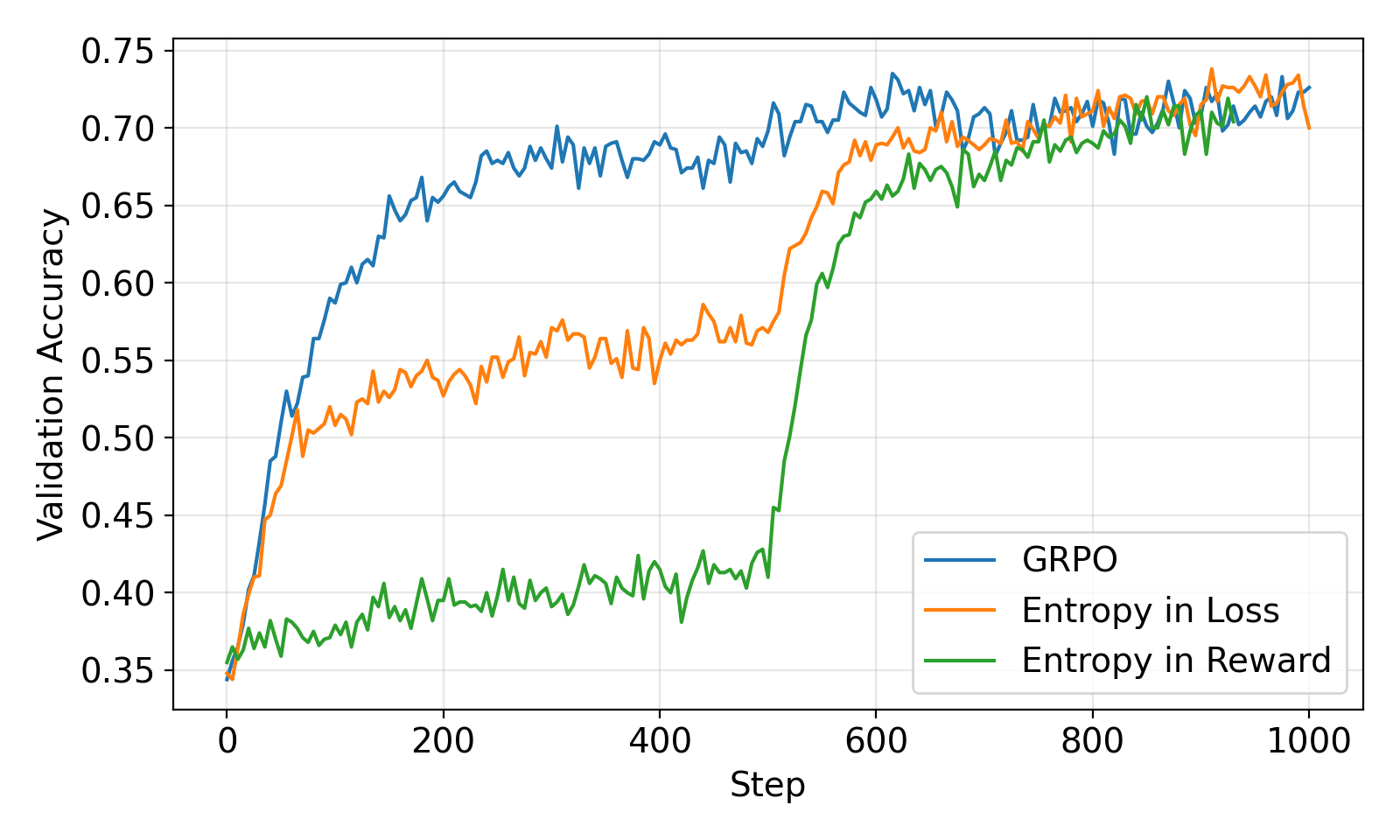}
        \caption{Validation accuracy over training steps.}
        \label{fig:multiple_intervention_acc}
    \end{subfigure}

    \caption{Entropy-intervention test: each method uses its own setting for the first 500 steps and then switches back to the same baseline for steps 500--1000.}
    \label{fig:multiple_intervention}
\end{figure}

From Figure~\ref{fig:multiple_intervention_entropy}, both interventions increase entropy, yet neither improves final performance. Figure~\ref{fig:multiple_intervention_acc} shows no post-intervention after-effect: the trajectories return toward the baseline once the interventions are removed, with early intervention-runs even lagging behind initially.  
These interventions act through different mechanisms and increase entropy by different amounts, yet neither produces a lasting performance gain. This suggests that entropy is a useful diagnostic correlate of convergence dynamics, but increasing it alone is not sufficient to improve the final plateau in our settings. This does not rule out a causal role for entropy in other interventions.

\begin{table}[ht]
\centering
\small
\setlength{\tabcolsep}{6pt}
\renewcommand{\arraystretch}{1.15}

\begin{tabularx}{\linewidth}{l c X}
\toprule

\makecell{\textbf{Entropy View}} 
& \textbf{Supported?} 
& \makecell{\textbf{What it suggests}} \\

\midrule

\textbf{A Sufficient Causal Lever} 
& \xmark 
& The tested early entropy interventions did not improve the final observed plateau. \\

\addlinespace[2pt]

\textbf{A Diagnostic Correlate} 
& \cmark 
& Entropy can indicate over-converging dynamics.\\
% \midrule
% \multicolumn{3}{c}{\textit{Entropy can be powerful, but only when used in the right way.}}\\ 
\bottomrule\end{tabularx}

\label{tab:entropy_scope}
\end{table}

\subsection{Sharpness Factor}
\label{sec:sources}

Consider the RL objective 
\[
J_{\mathrm{RL}}(\theta)
= \mathbb{E}\!\big[R(\gamma)\big]
- \beta\, D_{\mathrm{KL}}\!\big(\pi_\theta \,\|\, \pi_{\mathrm{ref}}\big),
\]
where $\pi_{\mathrm{ref}}$ is the reference policy,  
$R(\gamma)$ is the reward of a trajectory $\gamma$, 
and $\beta > 0$ penalizes deviation from $\pi_{\mathrm{ref}}$. 
Prior works often explain premature convergence by analyzing its local gradients. Here, we take a global view: this objective is equivalent to minimizing the \emph{reverse KL} between $\pi_\theta$ and the \textbf{target policy} $\pi^*$ (\S\ref{app:rl_forward}):
\begin{equation}
\label{eq:adv_diff}
\pi^*(\gamma)
\propto
\pi_{\mathrm{ref}}(\gamma)\,
\exp\!\left({R(\gamma)}/{\beta}\right),
\end{equation}

This view makes the effect of RL transparent: it pushes $\pi_\theta$ toward a \textit{reward-weighted} version of $\pi_{\mathrm{ref}}$. On a finite response space, as $\beta\to0^+$, this target concentrates on reward-maximizing responses within the support of $\pi_{\mathrm{ref}}$. The limit is deterministic only when the supported maximizer is unique; with ties, it retains the reference policy's relative weights among the maximizers.

Consider two responses $o^+$ and $o^-$ to a prompt $p$ that receive a higher reward $r^+$ and a lower one $r^-$, respectively. From Eq~\ref{eq:adv_diff}, the relative probability that $\pi^*$ assigns to $o^+$ over $o^-$ is exponentially controlled by the \textbf{(principled) sharpness factor} $\kappa$:
\begin{equation}
\frac{\pi^*(o^+\mid p)}{\pi^*(o^-\mid p)} = \frac{\pi_{\mathrm{ref}}(o^+\mid p)}{\pi_{\mathrm{ref}}(o^-\mid p)}\exp\!\left(\frac{r^+-r^-}{\beta}\right),
\qquad
\boxed{
\kappa:=\frac{r^+-r^-}{\beta},
}
\label{eq:ratio_a}
\end{equation}
For $\beta>0$, $\kappa$ controls the \emph{multiplicative amplification} of higher- versus lower-reward response odds relative to the reference policy. For example, $\beta=0.001$, $r^+=1$, and $r^-=0$ give $\kappa=1000$, so the target odds are the reference odds multiplied by $e^{1000}\approx1.97\times10^{434}$. This characterizes the idealized target, rather than the speed with which a parameterized optimizer reaches it. Strong concentration can reduce probability assigned to lower-reward responses, some of which may contain useful reasoning segments.

The target-policy calculation above describes a KL-regularized objective. Practical algorithms also use normalized advantages and importance-ratio clipping, so their realized updates need not equal that idealized target. For a chosen higher- and lower-advantage response pair, we define an operational \textbf{effective sharpness factor} as the change in log odds:
\begin{equation}
\boxed{\kappa_{\mathrm{eff}}(s;a^+,a^-)
:=\log\frac{\pi_{\mathrm{new}}(a^+\mid s)}{\pi_{\mathrm{new}}(a^-\mid s)}
-\log\frac{\rho(a^+\mid s)}{\rho(a^-\mid s)}.}
\label{eq:sharpness_factor_eff}
\end{equation}
Here $\rho$ is the policy before the update, and all four probabilities must be positive. In the reported measurements, each response is a full sequence: after updating a batch, we recompute log-probabilities for the same sampled responses and sum token log-probability changes relative to their stored rollout policy. For each prompt, we select one maximum-advantage and one minimum-advantage response and subtract their sequence log-probability changes. We skip groups with fewer than two responses or a nonpositive advantage gap, and report the mean across valid prompts. For an idealized KL mirror step with temperature $\tau>0$, this statistic equals $\Delta\tilde A(s)/\tau$ (Appendix~\ref{app:the_effective_kappa}). For a general update, $\tau_{\mathrm{eff}}:=\Delta\tilde A(s)/\kappa_{\mathrm{eff}}$ is only a pair-specific diagnostic when the denominator is nonzero; it need not be positive. PPO-style surrogate clipping alone does not establish an exact positive-temperature projection or a hard bound on this statistic.

Intuitively, $\kappa_{\text{eff}}$ measures how aggressively RL turns advantage gaps into log-odds gaps after a single update. Hence a larger $\kappa_{\text{eff}}$ implies more sharpening pressure on $\pi_{\theta}$. Loosely speaking, although larger $\kappa$ often indicates a more converging model that may give better performance, an ideal training trajectory should have large $\kappa$ mainly for genuine learning signals, and preserve room for lower-reward ones that may still contain useful segments to support better overall convergence.

% Later, in \S\ref{sec:reward_redesign}, we show that different algorithms induce different values of $\kappa$ and $\kappa_{\text{eff}}$, and we analyze how these differences affect training dynamics. This motivates our reward-budgeting method, which implicitly regulates $\kappa$---a factor our evidence suggests is more closely tied to causal effects---rather than explicitly controlling entropy, which appears to be primarily a correlate.

Later, in \S\ref{sec:reward_redesign}, we compare how different advantage constructions shape $\kappa$ and examine their implications for training dynamics. This motivates reward budgeting, which regulates sharpening pressure through reward allocation rather than explicit entropy control.

Generally speaking, entropy, $\kappa$ and $\kappa_{\text{eff}}$ differ in granularity and the use of response-quality signals. Entropy measures stepwise output randomness on training data and does {not} indicate response quality. $\kappa$ includes $r^{+,-}$ to distinguish good and bad responses, but it is a global metric shared across all steps. $\kappa_{\text{eff}}$ extends $\kappa$ to the step level, preserving quality distinction while providing finer discrimination.

% \subsubsection{Early Intervention for Sharpeness Factor}

% We repeat the early-intervention experiment previously conducted for entropy, this time varying the sharpness factor \(\kappa\) during the first 500 steps. All runs then switch to the same baseline \((r^+-r^-, \beta)=(5, 0.001)\). Although they eventually approach similar performance plateaus, they converge at different rates, with the effects of early \(\kappa\) interventions persisting for over 450 steps after the switch. In contrast, the effects of entropy interventions largely dissipate within approximately 100 steps. These results suggest that an early intervention does not necessarily determine the asymptotic endpoint under a sufficiently long common continuation, but it causally controls sharpening dynamics.

\subsubsection{Sharpness Factor Controlled Studies}
\label{sec:sharpness_factor_controlled_studies}

Compared with the tested entropy interventions in \S\ref{sec:entropy_is_not_the_cause}, changes to reward scale and KL strength produce clearer differences in finite-horizon training trajectories. We summarize these configurations through $\kappa$ below; the comparison does not isolate it as a unique causal mediator or determine an asymptotic endpoint.

\begin{wrapfigure}{r}{0.48\linewidth}
    \centering
    \vspace{-10pt}
    \includegraphics[width=\linewidth]{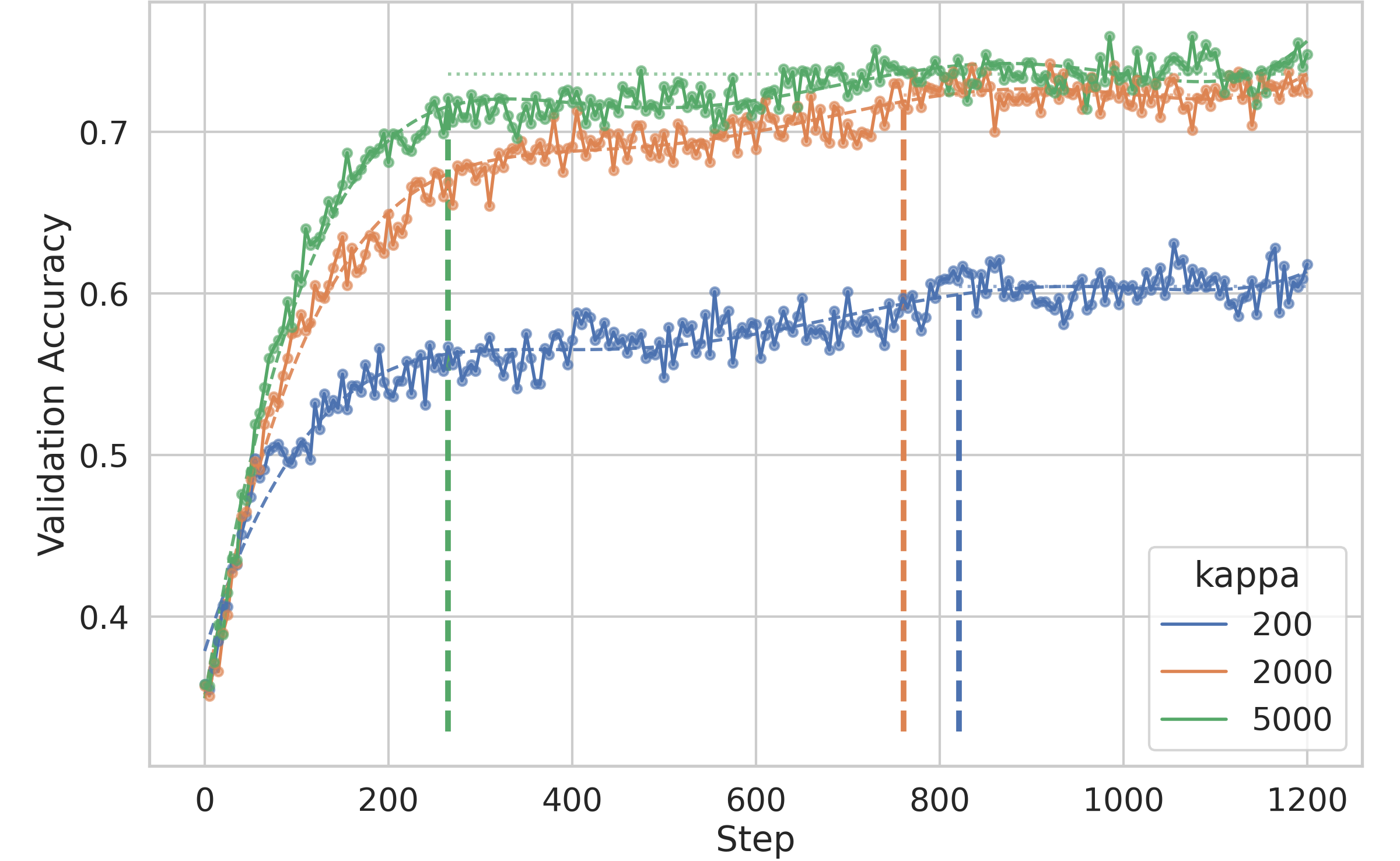}
    \caption{Validation accuracy as training progresses. The three vertical dashed
    lines mark the convergence step $t_{\text{conv}}$. Overall, smaller $\kappa$ results in
    later convergence but also slows training.}
    \label{fig:sharpness_factor}
    \vspace{-10pt}
\end{wrapfigure}

Specifically, we train Qwen3-8B~\citep{yang2025qwen3technicalreport} on DAPO-Math-17K~\citep{yu2025dapoopensourcellmreinforcement} with DRGRPO~\citep{liu2025understanding}, under three configurations: 
$(r^+-r^-, \beta) = (5, 0.001)$, $(2, 0.001)$, and $(1, 0.005)$, yielding decreasing sharpening pressures of $\kappa = 5000, 2000, 200$. To simplify the analysis, we use $\kappa$ instead of $\kappa_{\text{eff}}$, since the two serve as proxies for each other. We vary only $\kappa$ and the related quantities $r^+$, $r^-$, and $\beta$, while keeping other hyper-parameters fixed.

For each configuration, we define $t_\text{conv}$ as the first point where performance plateaus (\S\ref{app:convergence_step_detection}). We visualize $t_{\text{conv}}$ by vertical lines. In Figure~\ref{fig:sharpness_factor}, larger $\kappa$ indeed exhibits markedly earlier $t_\text{conv}$, but smaller $\kappa$ also yields lower final performance. This aligns with \S\ref{sec:sources}, where $\kappa$ (and $\kappa_{\text{eff}}$) are derived from sharpening pressure. Making $\kappa$ globally small can slow learning substantially \emph{without} guaranteeing higher final accuracy; see more discussion in \S\ref{sec:reward_redesign}.

Under the same set-up, we provide a complementary conditional association analysis in~\ref{app:deconfounding_regression}, showing that entropy is much less predictive of future accuracy once variation in $\kappa$ is controlled for.

% \begin{table}[ht]
% \centering
% \small
% \begin{tabular}{c|ccc}
% \toprule
% Config & $A^+-A^-$ & $\beta$ & $\kappa:=(A^+-A^-)/\beta$ \\
% \midrule
% 1 & 5 & 0.001 & 5000 \\
% 2 & 2 & 0.001 & 2000  \\
% 3 & 1 & 0.005  & 200   \\
% \bottomrule
% \end{tabular}
% \caption{{Training configurations for sharpness factor controlled studies.} Configs 1--3 represent a decreasing gradient of sharpness levels used to assess their effect on training dynamics.}
% \label{tab:sharpness_configs}
% \end{table}

\subsection{The Averaging Effect}
\label{sec:the_smoothing_trap}

When plotting entropy and accuracy, ``entropy collapse'' appears striking: accuracy plateaus as entropy drops. 
However, this observation is usually based on heavily averaged curves, such as 11 runs in \citep{cui2025entropymechanismreinforcementlearning} and 12 runs in our Figure~\ref{fig:entropy_collapse}. Although averaging reduces noise, it also introduces a subtle \textbf{statistical risk}: by smoothing away run-level heterogeneity, it can make weak or inconsistent trends appear visually robust. Therefore, averaged plots and individual runs can show markedly different trends. This distinction is particularly important for causal interpretation. The relationship between a strongly tied metric and accuracy should remain visible within individual runs, not only after averaging.

\begin{figure}[htbp]
    \centering

    \begin{subfigure}{0.49\linewidth}
        \centering
        \includegraphics[width=\linewidth]{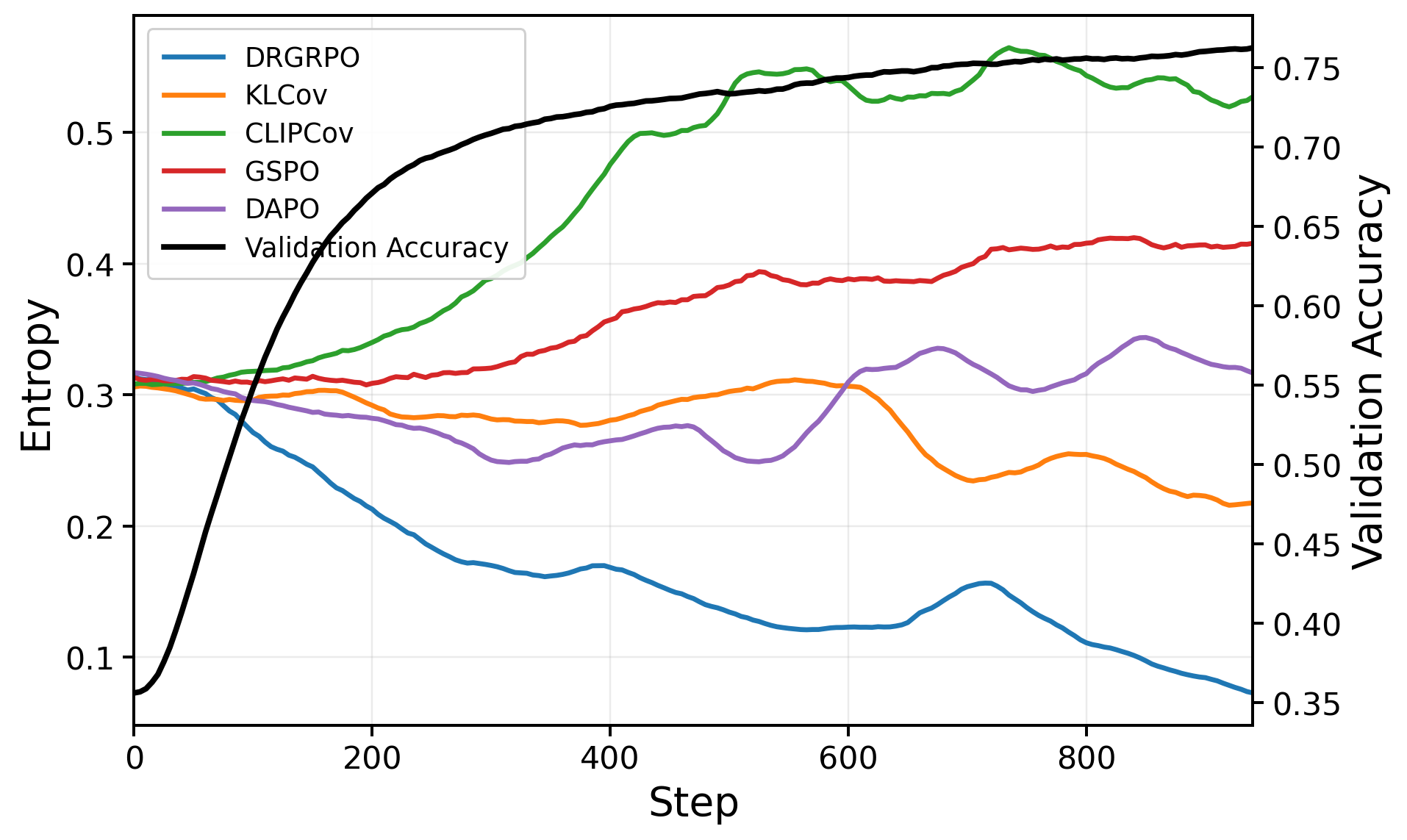}
        \caption{Entropy over training steps.}
        \label{fig:entropy_validation_accuracy_qwen}
    \end{subfigure}
    \hfill
    \begin{subfigure}{0.49\linewidth}
        \centering
        \includegraphics[width=\linewidth]{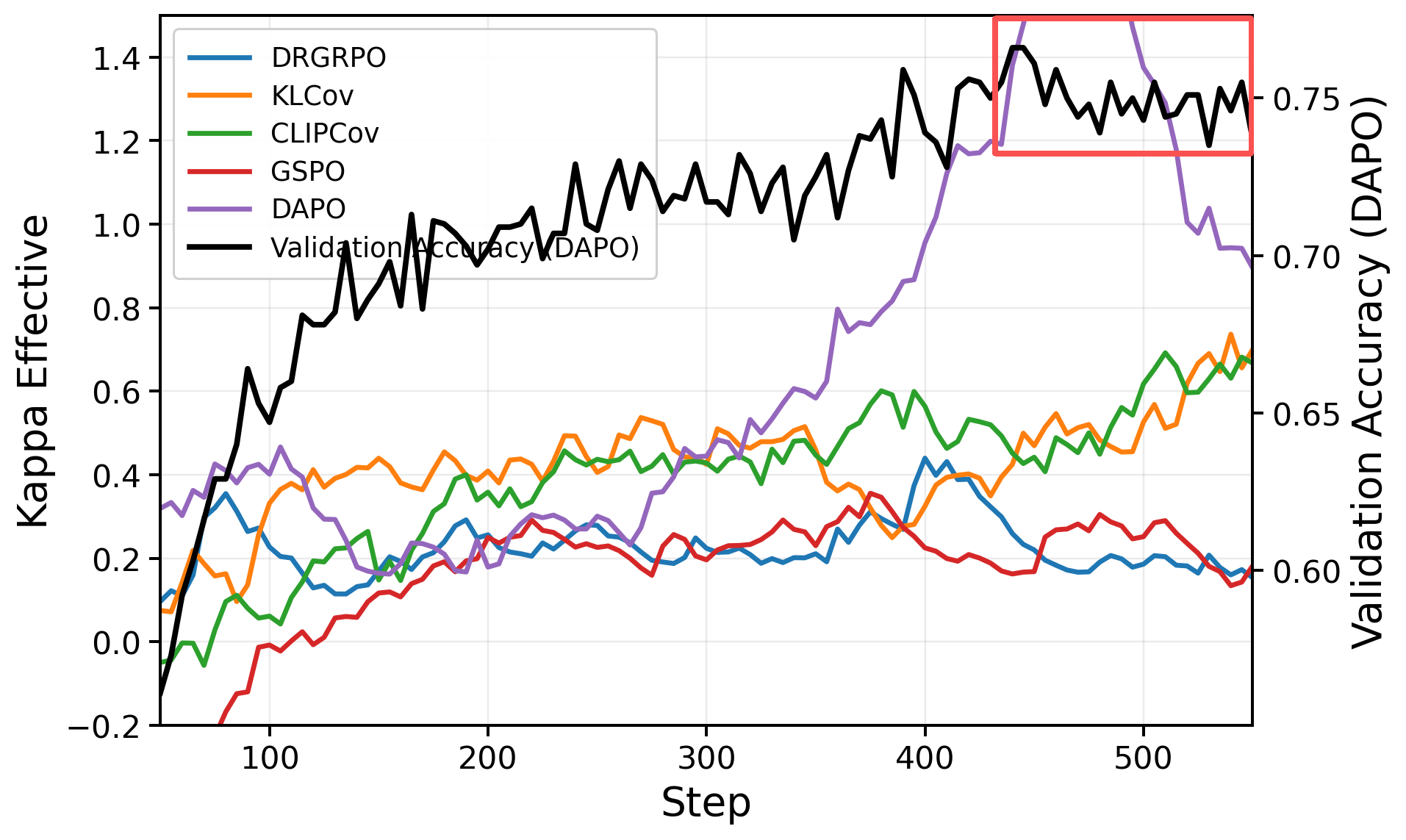}
        \caption{$\kappa_{\text{eff}}$ over training steps.}
        \label{fig:kappa_validation_accuracy_qwen}
    \end{subfigure}
    \caption{Training dynamics for DRGRPO, KLCov, CLIPCov, GSPO, and DAPO. The left axes show entropy in (a) and $\kappa_{\mathrm{eff}}$ in (b); the right axes show averaged validation accuracy in (a) and DAPO validation accuracy in (b).}
    \label{fig:entropy_vs_kappa_qwen}
\end{figure}

Figure~\ref{fig:entropy_vs_kappa_qwen} shows substantial between-method variation in the diagnostic trajectories. These plots provide descriptive comparisons and do not establish a general within-run relationship or a causal connection between entropy, effective sharpness, and accuracy.

% Under this criterion, entropy behaves more like a correlate: it is informative in averaged plots, but unstable as a run-level explanation. In other words, there is little run-level evidence that learning systematically plateaus when entropy collapses. $\kappa_{\text{eff}}$, by contrast, retains its alignment with accuracy within individual runs, making it a stronger candidate for a causal indicator. This supports our broader argument that $\kappa_{\text{eff}}$ is closer to the causal structure of training dynamics than entropy.

\section{Method}
\label{sec:garo}

We propose a plug-and-play advantage construction method, \textbf{reward budgeting}, to regulate RL training dynamics. Despite its simplicity, reward budgeting has a principled theoretical effect. First, under the conditions below, its unclipped score-function estimator follows a finite-$G$ concave objective. Second, it yields a $\kappa$ trajectory that is different from most existing RL methods which may better allocate training focus. Its empirical gains are further verified in \S\ref{sec:experiment}. 

Reward budgeting changes how advantages are constructed and can be used with a fixed-reference KL penalty. Section~\ref{sec:prior_never_fades} discusses the reference-free variant, which we evaluate separately in the Qwen3-8B GRPO ablation.

\subsection{Fixed Per-Prompt Reward Budgets}
\label{sec:reward_redesign}

\begin{algorithm}[htbp]
\small
\caption{Advantage Construction in Fixed Per-Prompt Reward Budgets}
\label{alg:garo}
\begin{enumerate}
    \item Initialize the policy $\pi_\theta$ and set the reference policy 
    $\pi_{\text{ref}} \leftarrow \pi_\theta$;  
    initialize training data $\mathcal{D}$ and the number of rollouts per prompt $G$.
    \item For each training step $t = 1, 2, \ldots$:
        \begin{enumerate}
            \item Sample $B$ prompts $\{p_i\}_{i=1}^B$ from $\mathcal{D}$.
            \item Generate $G$ rollouts 
            $\{o_i^{(j)}\}_{j=1}^G \sim \pi_\theta$ for each $p_i$.
            \item Assign raw rewards $r_i^{(j)}$ to each $o_i^{(j)}$.
            \item Compute per-prompt minimum $r_i^-:=\min_{1\le j\le G} r_i^{(j)}$.
            \item Subtract the minimum $g_i^{(j)}:=r_i^{(j)} - r_i^-$.
            \item If a prompt $p_i$ has $g_i^{(j)}=0$ for all $j\in\{1,\dots,G\}$, skip the prompt; otherwise:
            \begin{enumerate}
                \item Redesign rewards
                $
                R_i^{(j)} :=
                {g_i^{(j)}}/{\sum_{k=1}^{G} g_i^{(k)}}.
                $
                \item Compute $A_i^{(j)}:=R_i^{(j)}-\frac{1}{G}\sum_j R_i^{(j)}$.
            \end{enumerate}
        \end{enumerate}
\end{enumerate}
\end{algorithm}

Reward budgeting constructs advantages as follows: for each prompt $p$, let $\pi_{\theta}$ sample $G$ rollouts $\{o_i\}_{i=1}^G$ with raw rewards $\{r_i\}_{i=1}^G$. It first removes a per-group minimum ${r}^-:=\min_{i}r_i$ and keeps only the positive ``gaps” $g_i:=r_i-r^-$. If all $g_i=0$, it skips the update. Otherwise it redistributes a unit total reward across the rollouts, proportional to their gaps: $R(o_i):={g_i}/{\sum_{j=1}^G g_j}$
so that $\sum_i R(o_i) = 1$. Finally, it defines the advantage as 
$A_i:=R(o_i)-\frac{1}{G}\sum_iR(o_i)$
to center the normalized rewards within the group; see Algorithm~\ref{alg:garo}. The objective analysis below retains this centering and the group-skipping rule.

\paragraph{Global Optimization Objectives.}
To interpret fixed reward budgets, we first compare their idealized score-function update with the standard expected-reward policy gradient~\citep{NIPS1999_464d828b}, whose objective is 
\[
J_{\text{lin}}=\mathbb{E}_{p}\,\mathbb{E}_{y\sim\pi_\theta(\cdot|p)}[R(y)] .
\]
For binary rewards, a sampled group of $G$ rollouts has total raw reward $k$ when $k$ rollouts are correct. Reward budgeting instead assigns a fixed total redistributed reward of $1$ to each mixed group. All-correct and all-incorrect groups are skipped because their reward gaps are zero. Thus additional correct rollouts within a mixed group share the same budget; the exact calculation below shows how this produces diminishing returns as success probability increases.

For binary rewards and independent on-policy rollouts, let $p_{\mathrm{correct}}:=p_\theta(\mathrm{correct}\mid p)$. Appendix~\ref{app:oo_binary} shows that the expected centered score-function estimator of Algorithm~\ref{alg:garo}, before clipping and other implementation-specific modifications, is the gradient of
\begin{equation}
J_{G-1}=\mathbb{E}_{p\sim\mathcal{D}}\big[f_{G-1}(p_{\mathrm{correct}})\big],
\quad
f_{G-1}(x)=\sum_{m=1}^{G-1}\frac{1-(1-x)^m}{m}.
\label{eq:binary_objective}
\end{equation}
This objective is approximately linear up to scale near zero and gives diminishing returns as success probability grows. The result is exact for this estimator at $G=8$; only its logarithmic limit requires large $G$.

For general rewards, Appendix~\ref{app:oo_gap} gives sufficient conditions under which the actual gap-normalized estimator has the limiting expected direction associated with
\begin{equation}
J_{\mathrm{gap}}=\mathbb{E}_{p\sim\mathcal{D}}\left[\log\left(\mathbb{E}_{y\sim\pi_\theta}[R(y)\mid p]-a(p)\right)\right],
\label{eq:general_objective}
\end{equation}
where $a(p)$ is the finite, locally policy-independent essential infimum of the reward distribution. This is an asymptotic interpretation, not an exact finite-$G$ objective for arbitrary graded rewards.

The $O(1/G)$ mean-squared error and variance bounds in Appendix~\ref{app:oo_general} concern a separate idealized, non-shifted estimator under stronger assumptions; the resulting bias bound is $O(G^{-1/2})$. They neither establish a rate for the actual gap-normalized estimator nor guarantee approximation accuracy at $G=8$. Clipping, importance weighting, and token-level optimization further limit equivalence to the evaluated implementations.

% Intuitively, reward budgeting raises the probability of being correct, but gives diminishing returns to redundant successes, mitigating reward hacking and premature convergence.

\paragraph{Sharpness Factor.}
Although reward budgeting applies only a fixed per-prompt reward re-allocation, it turns out to \emph{implicitly} regulate $\kappa$ in a way that differs from existing RL methods. 
Specifically, we consider three representative forms of advantage construction: 

\begin{itemize}[leftmargin=1.5em, itemsep=2pt, topsep=2pt]
\item DRGRPO uses $A=r-\bar{r}$, giving $\kappa_{\mathrm{DR}}=1/\beta$;
\item GRPO, GSPO, DAPO, ClipCov and KLCov use $A=(r-\bar{r})/\sigma$, giving
$\kappa_{\mathrm{GRPO}}=\frac{1}{\beta\sigma}$;
\item Reward budgeting gives $\kappa_{RB}=\frac{1}{\beta k}$. 
\end{itemize}

\begin{wraptable}{r}{0.5\columnwidth}
\vspace{-10pt}
\centering
\scriptsize
\caption{Correct/incorrect samples' advantages in 2 decimal places across different constructions: Reward Budget, DrGRPO and GRPO (most RL methods fall into this type). $G$ is the group size and $k$ is the number of correct rollouts.}
\label{tab:advantage_no_garo}
\setlength{\tabcolsep}{2.2pt}
\renewcommand{\arraystretch}{0.92}
\resizebox{\linewidth}{!}{%
\begin{tabular}{c ccc ccc ccc}
\toprule
& \multicolumn{3}{c}{\textbf{Reward Budget}}
& \multicolumn{3}{c}{\textbf{DrGRPO}}
& \multicolumn{3}{c}{\textbf{GRPO (Most)}} \\
\cmidrule(lr){2-4}\cmidrule(lr){5-7}\cmidrule(lr){8-10}
\textbf{$k/G$}
& $+$ & $-$ & $\kappa$
& $+$ & $-$ & $\kappa$
& $+$ & $-$ & $\kappa$ \\
\midrule
1/8 & 0.88 & -0.13 & 1000 & 0.88 & -0.13 & 1000 & 2.65 & -0.38 & 3024 \\
2/8 & 0.38 & -0.13 & 500  & 0.75 & -0.25 & 1000 & 1.73 & -0.58 & 2309 \\
3/8 & 0.21 & -0.13 & 333  & 0.63 & -0.38 & 1000 & 1.29 & -0.78 & 2066 \\
4/8 & 0.13 & -0.13 & 250  & 0.50 & -0.50 & 1000 & 1.00 & -1.00 & 2000 \\
5/8 & 0.08 & -0.13 & 200  & 0.38 & -0.63 & 1000 & 0.78 & -1.29 & 2066 \\
6/8 & 0.04 & -0.13 & 167  & 0.25 & -0.75 & 1000 & 0.58 & -1.73 & 2309 \\
7/8 & 0.02 & -0.13 & 143  & 0.13 & -0.88 & 1000 & 0.38 & -2.65 & 3024 \\
\midrule
\multicolumn{1}{c}{$\kappa$ form}
& \multicolumn{3}{c}{${1}/{\beta k}$}
& \multicolumn{3}{c}{${1}/{\beta}$}
& \multicolumn{3}{c}{${1}/{\beta\sigma}$} \\
\bottomrule
\end{tabular}%
}
\vspace{-10pt}
\end{wraptable}

These $\kappa$ values show how different methods pose sharpening pressure. First, DRGRPO applies uniform sharpening pressure, which is stable but may risk under-convergence. Second, GRPO-style advantages maximize $\kappa$ at extreme $k\in\{1,G-1\}$, so the strongest sharpening occurs when a batch is almost all correct or almost all incorrect. While this can speed convergence, when a batch is almost all correct, such sharpening is less necessary, because the model is already near perfect. Finally, reward budgeting gives $\kappa=\frac{1}{k\beta}$, assigning stronger sharpening to harder prompts.
Thus, ``easy'' prompts do not keep generating increasingly sharp updates, reducing the incentive to overfit to cheaply repeatable templates. 

This sharpness view complements the scoped objective analysis: reward budgeting adjusts the advantage gaps governing idealized sharpening. This provides a mechanism consistent with the observed gains, while the diagnostics do not isolate the causal contribution of each implementation component.

\paragraph{Stepwise KL.} Fixed reward budgeting also permits a local surrogate interpretation. For a fixed non-constant sampled group and detached budget weights, consider
\[\nabla_\theta\sum_{i=1}^G R(o_i)\log\pi_\theta(o_i\mid p).\]
The nonnegative weights sum to one and define an empirical target distribution, combining weights for duplicate responses. Maximizing this fixed-sample weighted log-likelihood minimizes the KL divergence from that empirical target to the policy, up to a constant. This surrogate omits group-relative centering and clipping; it is not the full implemented objective.

\subsection{The Prior Never Fades} 
\label{sec:prior_never_fades}

Usually $\pi_{\text{ref}}$ is introduced to promote stability by penalizing the current policy for drifting too far, but it in fact plays a deeper role: $\pi^*\propto\pi_{\text{ref}}\exp(\frac{1}{\beta}A)$ permanently depends on $\pi_{\text{ref}}$, so $\pi_{\text{ref}}$ shapes what the policy is trained to approximate, and its impact never vanishes no matter how long we train.

The reference-free variant removes the KL penalty while retaining importance-ratio clipping. Removing the fixed-reference penalty changes the unconstrained target to reward maximization, whose optima need not be deterministic when rewards tie. Clipping modifies the incentive for large sampled importance-ratio changes around the previous policy; it does not impose a hard global trust region or guarantee a bound on $\kappa_{\mathrm{eff}}$. The empirical effects of removing KL and adding reward budgeting are distinguished in the ablation study.

\section{Experiment}

% reward budgets improves late-stage learning without increasing entropy

% 控制 kappa > 控制熵

% 多 seed 和方差

% related work

% 成本

% kappa_eff 监控曲线

% hyper-params: importance clipping epsilon, etc.

\begin{table}[t]
\centering
\caption{Benchmark scores (\%) from eight responses per prompt at temperature $0.7$: mean accuracy for Qwen3-8B and pass@8 for Llama3.2-3B-Instruct. Shaded rows show each baseline's percentage-point change with reward budgeting (RB); GRPO retains reference-policy KL.}
\label{tab:main_results}
\begin{tabular}{lcccccc}
\toprule
Method & AMC23 & AIME24 & AIME25 & MATH500 & Olympiad (Physics) & Average \\
\midrule
\multicolumn{7}{c}{Qwen3-8B} \\
\midrule
GRPO        & 82.50 & 26.67 & 43.33 & 95.60 & 51.27 & 59.87 \\
\rowcolor{oursgray}
+ RB        & \textcolor{gainGreen}{+10.00} & \textcolor{gainGreen}{+10.00} & \textcolor{gainGreen}{+10.00} & \textcolor{gainGreen}{+1.20} & \textcolor{gainGreen}{+2.12} & \textcolor{gainGreen}{+6.66} \\
DrGRPO      & 82.50 & 30.00 & 43.33 & 97.20 & 47.46 & 60.10 \\
\rowcolor{oursgray}
+ RB        & \textcolor{gainGreen}{+5.00} & \textcolor{gainGreen}{+6.67} & \textcolor{gainGreen}{+3.33} & \textcolor{red}{-0.20} & \textcolor{gainGreen}{+1.27} & \textcolor{gainGreen}{+3.21} \\
GSPO        & 90.00 & 26.67 & 36.67 & 96.80 & 47.04 & 59.44 \\
\rowcolor{oursgray}
+ RB        & \textcolor{gainGreen}{+5.00} & \textcolor{gainGreen}{+6.67} & \textcolor{gainGreen}{+6.67} & \textcolor{gainGreen}{+1.00} & \textcolor{gainGreen}{+1.69} & \textcolor{gainGreen}{+4.21} \\
DAPO        & 87.50 & 36.67 & 40.00 & 96.40 & 49.15 & 61.94 \\
\rowcolor{oursgray}
+ RB        & \textcolor{gainGreen}{+7.50} & \textcolor{gainGreen}{+6.67} & \textcolor{gainGreen}{+6.67} & \textcolor{gainGreen}{+1.60} & \textcolor{gainGreen}{+0.85} & \textcolor{gainGreen}{+4.66} \\
ClipCov     & 85.00 & 26.67 & 33.33 & 96.60 & 48.31 & 57.98 \\
\rowcolor{oursgray}
+ RB        & \textcolor{gainGreen}{+2.50} & \textcolor{gainGreen}{+10.00} & \textcolor{gainGreen}{+6.67} & \textcolor{gainGreen}{+1.40} & \textcolor{gainGreen}{+1.27} & \textcolor{gainGreen}{+4.37} \\
KLCov       & 82.50 & 33.33 & 43.33 & 96.80 & 49.58 & 61.11 \\
\rowcolor{oursgray}
+ RB        & \textcolor{gainGreen}{+5.00} & \textcolor{gainGreen}{+10.00} & \textcolor{gainGreen}{+6.67} & \textcolor{gainGreen}{+0.80} & \textcolor{gainGreen}{+2.12} & \textcolor{gainGreen}{+4.92} \\
% \midrule
% \rowcolor{oursgray}
% GARO (ours) & \textbf{95.40 } & 85.00 & \textbf{50.00 } & \textbf{53.70 } & \textbf{29.12 } & \textbf{60.73} \\
\midrule
\multicolumn{7}{c}{Llama3.2-3B-Instruct} \\
\midrule
GRPO        & 7.50  & 3.33  & 0.00  & 67.60 & 20.31 & 19.75 \\
\rowcolor{oursgray}
+ RB        & \textcolor{gainGreen}{+12.50} & \textcolor{blue}{+0.00} & \textcolor{gainGreen}{+3.33} & \textcolor{gainGreen}{+3.20} & \textcolor{gainGreen}{+1.53} & \textcolor{gainGreen}{+4.11} \\
DrGRPO      & 30.00 & 10.00 & 0.00  & 68.60 & 14.94 & 24.71 \\
\rowcolor{oursgray}
+ RB        & \textcolor{gainGreen}{+12.50} & \textcolor{gainGreen}{+6.67} & \textcolor{gainGreen}{+6.67} & \textcolor{gainGreen}{+2.00} & \textcolor{gainGreen}{+1.15} & \textcolor{gainGreen}{+5.80} \\
GSPO        & 22.50 & 3.33  & 0.00  & 72.20 & 14.83  & 22.57 \\
\rowcolor{oursgray}
+ RB        & \textcolor{gainGreen}{+7.50}  & \textcolor{gainGreen}{+3.33} & \textcolor{gainGreen}{+3.33} & \textcolor{gainGreen}{+5.00} & \textcolor{gainGreen}{+1.27} & \textcolor{gainGreen}{+4.01} \\
DAPO        & 17.50 & 6.67  & 3.33  & 70.40 & 21.84 & 23.95 \\
\rowcolor{oursgray}
+ RB        & \textcolor{gainGreen}{+7.50}  & \textcolor{gainGreen}{+3.33} & \textcolor{blue}{+0.00} & \textcolor{gainGreen}{+2.20} & \textcolor{gainGreen}{+1.47} & \textcolor{gainGreen}{+2.90} \\
ClipCov     & 22.50 & 10.00 & 0.00  & 69.80 & 13.98 & 23.26 \\
\rowcolor{oursgray}
+ RB        & \textcolor{gainGreen}{+12.50} & \textcolor{gainGreen}{+3.33} & \textcolor{gainGreen}{+3.33} & \textcolor{gainGreen}{+3.40} & \textcolor{gainGreen}{+1.27} & \textcolor{gainGreen}{+4.77} \\
KLCov       & 30.00  & 10.00 & 0.00  & 71.60 & 13.98  & 25.12  \\
\rowcolor{oursgray}
+ RB        & \textcolor{gainGreen}{+10.0} & \textcolor{gainGreen}{+6.67} & \textcolor{gainGreen}{+3.33} & \textcolor{gainGreen}{+2.40} & \textcolor{gainGreen}{+1.69} & \textcolor{gainGreen}{+4.82} \\
\bottomrule
\end{tabular}
\end{table}

\label{sec:experiment}

% (1) 收敛步数 t_{conv}：最近 M 次评估提升 < \epsilon。
% (2) 早期学习速率：前 20% steps 的斜率
% (3) 最终性能：peak 或固定 budget 下的最后 K steps 平均
% (4) 更新强度 KL (有些算法没有 KL term)
% (5) rollout 权重集中度: garo rollout weight - gini/entropy/n_eff; std(A)
% (6) 优势差距统计：每 step 的 kappa
% (7) policy entropy
% (8) clip ratio
% (9) 学习原本不会的知识

\subsection{Setup}
\label{sec:setup}

% We validate GARO in mathematical reasoning, one of the most active recent application domains of RL. We leave evaluation on other data domains to future work.

\textbf{Data.} We train on DAPO-Math-17K~\citep{yu2025dapoopensourcellmreinforcement} and evaluate on five benchmarks: AMC23~\citep{aimo_amc_subset_2024}, 
AIME24~\citep{h4_aime2024},
AIME25~\citep{aime25}, 
MATH500~\citep{MATH},
and OlympiadBench (Physics)~\citep{he-etal-2024-olympiadbench}. 
% The first four benchmarks are mathematical reasoning tasks, while Olympiad (Physics) is used to illustrate out-of-distribution (OOD) generalization.

\textbf{Models.} We train (1) Llama3.2-3B-Instruct~\citep{meta2024llama32_3b_instruct} and (2) Qwen3-8B~\citep{yang2025qwen3technicalreport}.
% They differ in scale and family, illustrating the generalization ability of our method.

\textbf{Hyper-parameters.} 
We sample eight rollouts per prompt and use 32 prompts per update. We train each method until its validation accuracy reaches a plateau. 
During evaluation, we set the maximum response length to $8192$ tokens and generate $8$ responses per prompt at temperature$=0.7$. We report their average accuracy for Qwen3-8B and pass@8 for Llama3.2-3B-Instruct (see detailed settings in \S\ref{app:hyper}). 
% For AMC23 and AIME25, whose ground-truth answers are integers, we use exact match; for other benchmarks, we use an LLM-as-Judge protocol with Qwen3-235B-Instruct. 

\textbf{Baselines.} We apply our method to six RL methods in two types:
(1) Four Representative RL Methods: GRPO~\citep{shao2024deepseekmathpushinglimitsmathematical}, DrGRPO~\citep{liu2025understanding}, GSPO~\citep{zheng2025groupsequencepolicyoptimization}, and DAPO~\citep{yu2025dapoopensourcellmreinforcement}; (2) Two Entropy-Based RL Methods: ClipCov and KLCov \citep{cui2025entropymechanismreinforcementlearning}, which are entropy-based variants of DAPO.

% \begin{itemize}[leftmargin=1.5em, itemsep=2pt, topsep=2pt]
% \item Four Representative RL Methods: GRPO~\citep{shao2024deepseekmathpushinglimitsmathematical}, DrGRPO~\citep{liu2025understanding}, GSPO~\citep{zheng2025groupsequencepolicyoptimization}, and DAPO~\citep{yu2025dapoopensourcellmreinforcement};
% \item Two Entropy-Based RL Methods: ClipCov and KLCov \citep{cui2025entropymechanismreinforcementlearning}. They are partial variants of DAPO, which aim at increasing entropy by suppressing high-covariance tokens via either clipping or a KL penalty, respectively.
% \end{itemize}

\subsection{Key Insights}
\label{sec:main_results}

\paragraph{More robust gains than entropy control.} Table~\ref{tab:main_results} includes two remedies: entropy (ClipCov and KLCov as entropy-based variants of DAPO) and our method. However, within each dataset, results of entropy-control methods are mixed. On average, on Qwen3-8B, DAPO even outperforms its variants (DAPO 61.94 vs. ClipCov 57.98 vs. KLCov 61.11); on Llama3.2-3B, KLCov marginally outperforms DAPO (25.12 vs. 23.95) while ClipCov slightly under-performs (23.26 vs. 23.95). In contrast, our method consistently improves in almost all cases.

% \paragraph{Generalization to out-of-distribution tasks.} On the OOD Olympiad (Physics) set, our method also achieves consistent improvement (+1.55 on Qwen3-8B and +1.40 on Llama-3.2-3B-instruct on average), yet entropy control tends to degrade. This  distinction suggests that the observed gains of our method may not come from narrow in-distribution tuning but from more robust learning.

\paragraph{Impact of advantage construction on $\kappa$-dynamics.} Apart from reward budgeting, the RL baselines in Table~\ref{tab:main_results} construct advantages in mainly two ways: DRGRPO uses $A=r-\bar{r}$, while others use $A=(r-\bar{r})/\sigma$ (in fact, DAPO and its variants also includes length-based reward, but its magnitude is typically much smaller than $r$, thus we omit this for simplicity). These different advantage constructions then induce different $\kappa$-trajectories (\S\ref{sec:reward_redesign}). Specifically, DRGRPO tends to exhibit slightly lower $\kappa$, thus gentler sharpening pressure. Comparing DRGRPO to the others, we observe that its performance remains in the top tier. This agrees with our earlier discussion: well-regulated $\kappa$ can deliver better final convergence, though our method continues to improve even in this case.

\subsection{Training Dynamics}

\begin{figure}[htbp]
    \centering
    \begin{subfigure}{0.49\linewidth}
        \centering
        \includegraphics[width=\linewidth]{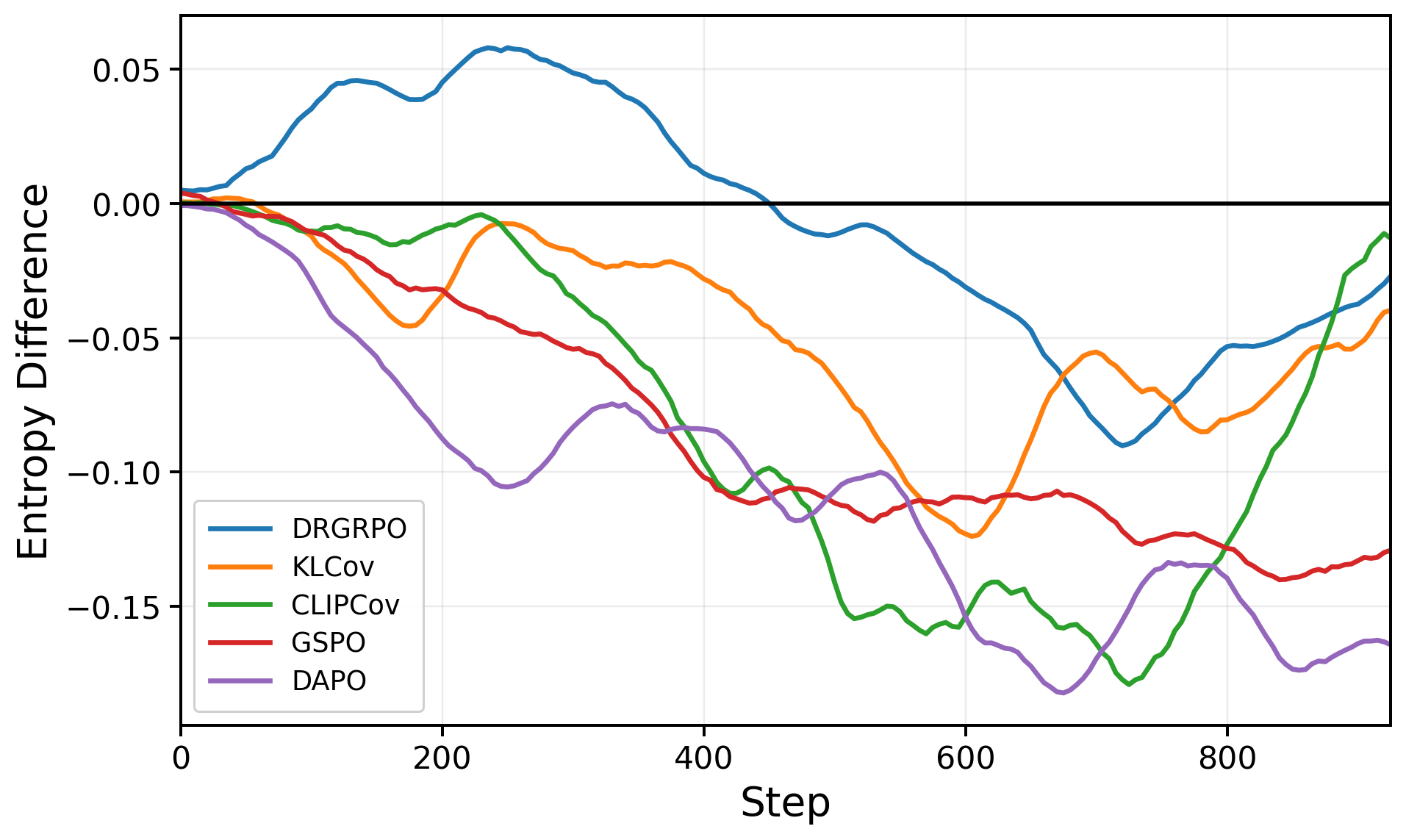}
        \caption{Entropy changes after applying our method}
        \label{fig:entropy_difference_qwen_rb_minus_qwen_with_zero_line}
    \end{subfigure}
    \hfill
    \begin{subfigure}{0.49\linewidth}
        \centering
        \includegraphics[width=\linewidth]{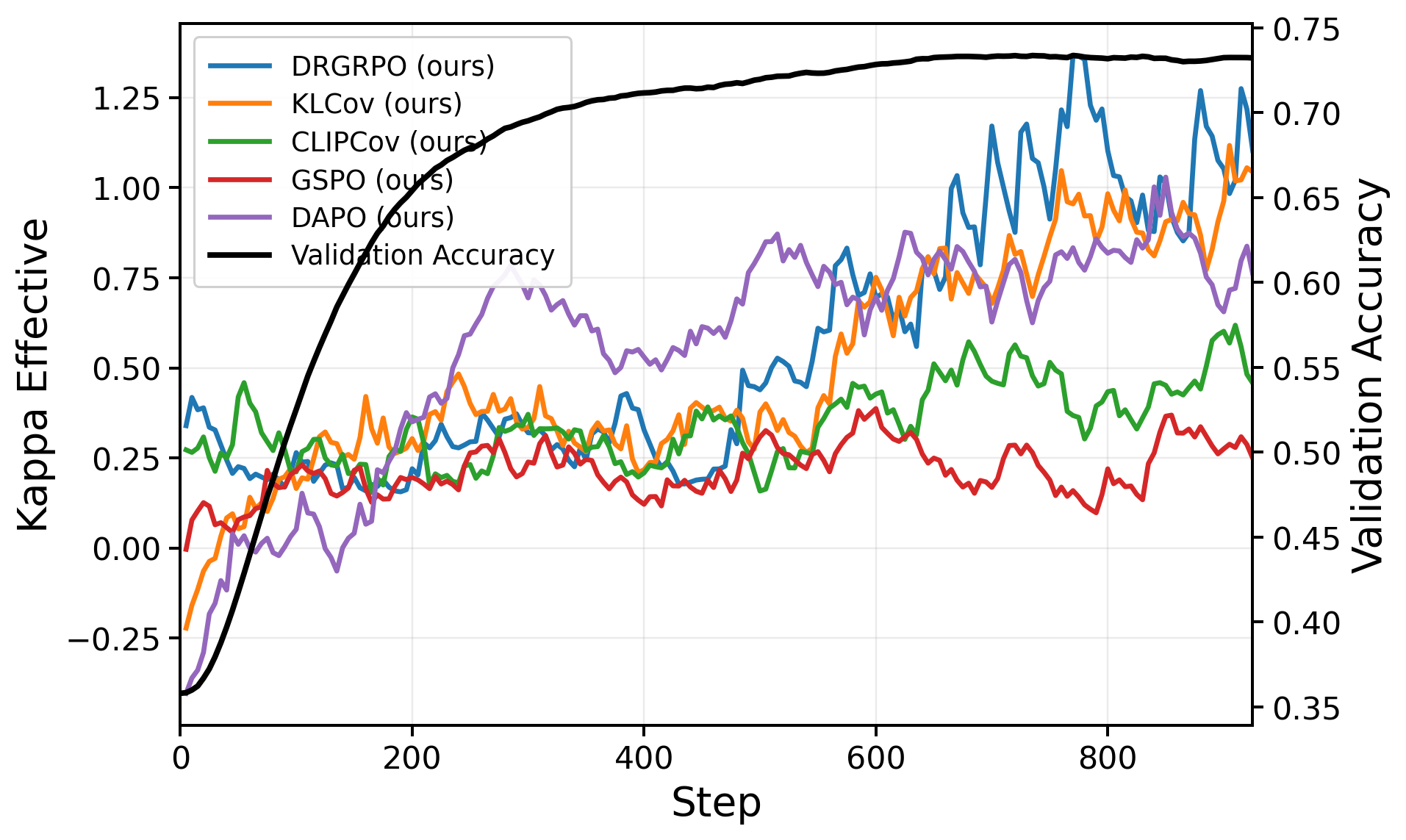}
        \caption{$\kappa_{\text{eff}}$ vs validation accuracy}
        \label{fig:kappa_validation_accuracy_qwen_rb}
    \end{subfigure}
    \caption{Reward budgeting on Qwen3-8B: (a) training-entropy differences from each baseline, mostly below zero; (b) effective sharpness and mean validation accuracy across methods (black curve). These descriptive trajectories do not establish the cause of the accuracy gains.}
    \label{fig:difference}
\end{figure}

From Figure~\ref{fig:entropy_difference_qwen_rb_minus_qwen_with_zero_line}, after applying our method, entropy generally decreases, since most curves remain below the black zero line. This indicates that higher entropy is not necessary for improving final performance, and that our method does not achieve its gains by implicitly boosting entropy. 
% In fact, this behaviour is expected: entropy measures the model's randomness on training data, so lower entropy can be interpreted as greater confidence and better convergence on the training distribution.
Comparing Figures~\ref{fig:kappa_validation_accuracy_qwen_rb} and~\ref{fig:kappa_validation_accuracy_qwen}, our method also regulates the $\kappa_{\text{eff}}$ trajectories with noticeably smaller oscillations, which is consistent with our discussion in \S\ref{sec:reward_redesign}. These trajectories accompany accuracy gains without a sustained increase in entropy.

% \subsection{Ablation and Practicality}

\subsection{Ablation}

The Qwen3-8B GRPO comparison in Table~\ref{tab:main_results} retains the reference-policy KL penalty both with and without reward budgeting. To assess how this penalty affects the result, Table~\ref{tab:kl_rb_ablation} crosses reward budgeting with KL retention or removal.

\begin{table}[t]
\centering
\caption{Qwen3-8B GRPO ablation of reference-policy KL and reward budgeting (RB); the first two rows give the absolute scores in Table~\ref{tab:main_results}.}
\label{tab:kl_rb_ablation}
\begin{tabular}{cccccccc}
\toprule
KL & RB & AMC23 & AIME24 & AIME25 & MATH500 & Olympiad (Physics) & Average \\
\midrule
$\checkmark$ & $\times$ & 82.50 & 26.67 & 43.33 & 95.60 & 51.27 & 59.87 \\
\rowcolor{oursgray}
$\checkmark$ & $\checkmark$ & 92.50 & 36.67 & 53.33 & 96.80 & 53.39 & \textbf{66.53} \\
$\times$ & $\times$ & 85.00 & 30.00 & 40.00 & 96.00 & 50.35 & 60.27 \\
$\times$ & $\checkmark$ & 95.00 & 36.67 & 46.67 & 97.00 & 52.89 & 65.65 \\
\bottomrule
\end{tabular}
\end{table}

Removing KL alone changes the average score from 59.87\% to 60.27\%, with mixed benchmark-level effects (Table~\ref{tab:kl_rb_ablation}). Reward budgeting improves the average by $+6.66$ percentage points with KL and $+5.38$ without KL, improving every evaluated benchmark in both settings. Thus, the gains are primarily attributable to reward budgeting rather than KL removal.

\section{Conclusion}

We study \emph{premature convergence} in RL and suggest that aggregate entropy may not fully characterize training dynamics. Instead, RL’s implicit target \(\pi^*\) reveals a built-in sharpening pressure, which is captured with the sharpness factor \(\kappa\) (and $\kappa_{\text{eff}}$). Motivated by this, we propose reward-budgeting to improve final performance without explicit entropy control. Appendix~\ref{sec:limitations} discusses limitations of these findings. 
% Overall, our findings suggest that mitigating premature convergence may require looking beyond direct entropy control and instead regulating training dynamics.

\section*{Declaration of LLM Usage}
Appendix~\ref{app:hyper} identifies Qwen3-235B-Instruct as the mathematical-equivalence judge and provides its system instruction and user-prompt template. LLM tools also assisted with checking mathematical arguments, refining the proof exposition, and editing the manuscript. The authors take responsibility for the correctness and originality of all mathematical arguments, experimental results, text, and references.

\bibliographystyle{plainnat}
\bibliography{references}

\clearpage
\appendix

\section{Proofs}

\subsection{The Effective Kappa}
\label{app:the_effective_kappa}

\begin{lemma}[Effective sharpness for a KL mirror step]
Fix a state $s$, a finite action space, a full-support policy $\rho$, detached advantages $\tilde A(s,a)$, and $\tau>0$. For the KL-regularized projection below and the log-odds definition in Eq.~\ref{eq:sharpness_factor_eff},
\[
\kappa_{\mathrm{eff}}(s;a^+,a^-)
=\frac{\tilde A(s,a^+)-\tilde A(s,a^-)}{\tau}.
\]
This identity describes the idealized projection, not an exact representation of arbitrary clipped policy updates.
\end{lemma}

\begin{proof}
Fix a prompt/state $s$ and write $\pi(a)\equiv \pi(a\mid s)$, $\rho(a)\equiv \rho(a\mid s)$, and $\tilde A(a)\equiv \tilde A(s,a)$.
Consider the KL-regularized projection (mirror descent) update
\begin{equation}
\begin{aligned}
\pi_{\mathrm{new}}
&\in \arg\max_{\pi}\Bigg\{
\sum_{a}\pi(a)\,\tilde A(a)
-\tau\, D_{\mathrm{KL}}(\pi\|\rho)
\Bigg\}
\end{aligned}
\end{equation}
such that
\begin{equation}
\begin{aligned}
\sum_a \pi(a)=1,
D_{\mathrm{KL}}(\pi\|\rho)
&:=\sum_a \pi(a)\log\frac{\pi(a)}{\rho(a)}.
\end{aligned}
\end{equation}
Introduce the Lagrangian
\begin{equation}
\begin{aligned}
\mathcal{L}(\pi,\lambda)
&=
\sum_a \pi(a)\tilde A(a)
-\tau \sum_a \pi(a)\log\frac{\pi(a)}{\rho(a)}
+\lambda\Big(\sum_a \pi(a)-1\Big).
\end{aligned}
\end{equation}
Taking derivatives and setting them to zero, for each $a$ we have
\begin{equation}
\begin{aligned}
0
=
\frac{\partial \mathcal{L}}{\partial \pi(a)}
&=
\tilde A(a)
-\tau\Big(\log\frac{\pi(a)}{\rho(a)}+1\Big)
+\lambda,
\\
\log\frac{\pi(a)}{\rho(a)}
&=
\frac{\tilde A(a)}{\tau}+\frac{\lambda-\tau}{\tau},
\\
\pi(a)
&=
\rho(a)\exp\!\Big(\frac{\tilde A(a)}{\tau}\Big)\cdot
\exp\!\Big(\frac{\lambda-\tau}{\tau}\Big).
\end{aligned}
\end{equation}
Let $C:=\exp((\lambda-\tau)/\tau)$, which does not depend on $a$. Enforcing normalization yields
\begin{equation}
\begin{aligned}
1=\sum_a \pi(a)
&=
C\sum_a \rho(a)\exp\!\Big(\frac{\tilde A(a)}{\tau}\Big)
\\
\Longrightarrow\quad
C
&=
\Bigg(\sum_{a'} \rho(a')\exp\!\Big(\frac{\tilde A(a')}{\tau}\Big)\Bigg)^{-1}.
\end{aligned}
\end{equation}
Therefore,
\begin{equation}
\begin{aligned}
\pi_{\mathrm{new}}(a\mid s)
&=
\frac{\rho(a\mid s)\exp(\tilde A(s,a)/\tau)}
{\sum_{a'}\rho(a'\mid s)\exp(\tilde A(s,a')/\tau)}.
\end{aligned}
\end{equation}
For any two actions/outputs $a^+$ and $a^-$, the normalizer cancels:
\begin{equation}
\begin{aligned}
\frac{\pi_{\mathrm{new}}(a^+\mid s)}{\pi_{\mathrm{new}}(a^-\mid s)}
&=
\frac{\rho(a^+\mid s)}{\rho(a^-\mid s)}
\exp\!\Big(\frac{\tilde A(s,a^+)-\tilde A(s,a^-)}{\tau}\Big),
\end{aligned}
\end{equation}
\begin{equation}
\begin{aligned}
\log\frac{\pi_{\mathrm{new}}(a^+\mid s)}{\pi_{\mathrm{new}}(a^-\mid s)}
-
\log\frac{\rho(a^+\mid s)}{\rho(a^-\mid s)}
&=
\frac{\tilde A(s,a^+)-\tilde A(s,a^-)}{\tau}.
\end{aligned}
\end{equation}
Defining $\Delta\tilde A(s):=\tilde A(s,a^+)-\tilde A(s,a^-)$ and
\begin{equation}
\begin{aligned}
\kappa_{\mathrm{eff}}(s;a^+,a^-)
:=
\log\frac{\pi_{\mathrm{new}}(a^+\mid s)}{\pi_{\mathrm{new}}(a^-\mid s)}
-
\log\frac{\rho(a^+\mid s)}{\rho(a^-\mid s)},
\end{aligned}
\end{equation}
we obtain the claimed identity
\begin{equation}
\begin{aligned}
\kappa_{\mathrm{eff}}(s;a^+,a^-)=\frac{\Delta\tilde A(s)}{\tau}
\quad (\text{and }\tau_{\mathrm{eff}}=\tau\text{ when }\Delta\tilde A(s)\ne0).
\end{aligned}
\end{equation}
\end{proof}

\subsection{Scope of the Objective Analysis}

Algorithm 1 uses a gap-normalized reward-budget estimator: it first subtracts the
within-group minimum reward, normalizes the remaining positive gaps to a fixed
per-prompt budget, and skips groups whose gaps are all zero. This differs from
the simpler reward-normalized estimator that directly uses \(R_i / \sum_j R_j\).

We therefore separate the objective analysis into three cases. First, in
Appendix~\ref{app:oo_binary}, we derive the exact finite-\(G\) objective induced by Algorithm 1
in the binary-reward setting. This is the setting most closely aligned with our
main RLVR experiments. Second, in Appendix~\ref{app:oo_general}, we analyze an idealized
non-shifted reward-normalized estimator, which gives the clean
\(\nabla \log \mathbb E[R]\) interpretation. This result should be understood
as an intuition for why fixed reward budgets induce diminishing returns, rather
than as the exact finite-\(G\) objective of Algorithm 1. Third, in Appendix~\ref{app:oo_gap},
we discuss the actual gap-normalized estimator under general rewards and show
that, under additional support assumptions, it asymptotically corresponds to a
shifted log-mean objective.

\subsection{Exact Objective of Algorithm 1 under Binary Rewards}
\label{app:oo_binary}

We first analyze the estimator exactly induced by Algorithm 1 in the binary
reward setting. This differs slightly from the simpler \(c_i/k\) estimator,
because Algorithm 1 subtracts the group minimum reward and skips groups whose
post-shift gaps are all zero.

\begin{assumption}[Binary reward setting]
For a prompt \(x\), sample \(G\) i.i.d. rollouts
\(y_1,\ldots,y_G \sim \pi_\theta(\cdot|x)\). Let
\(c(y)\in\{0,1\}\) denote correctness,
\(c_i := c(y_i)\), and
\[
k := \sum_{i=1}^G c_i,\qquad
p_\theta(x) := \Pr_{y\sim\pi_\theta(\cdot|x)}(c(y)=1).
\]
Assume $G\ge2$, $0<p_\theta(x)<1$, a correctness rule with no explicit $\theta$ dependence, locally parameter-independent policy support, and the integrability and differentiation conditions needed for the score-function identities below.
\end{assumption}

Under Algorithm 1, there are three cases:
\[
k=0: \text{ all rollouts are incorrect and the group is skipped,}
\]
\[
1\le k\le G-1: \text{ the group is mixed and contributes an update,}
\]
\[
k=G: \text{ all rollouts are correct and the group is skipped.}
\]
The last case is important: after subtracting the group minimum, all gaps become
zero, so the group contains no within-group preference contrast.

For \(1\le k\le G-1\), the shifted gaps are \(g_i=c_i\), the normalized
budgeted rewards are \(R_i=c_i/k\), and Algorithm 1 uses the group-relative
advantage
\[
A_i=\frac{c_i}{k}-\frac{1}{G}.
\]
Thus the exact estimator is
\[
\hat g_{\mathrm{RB}}(x)
=
\mathbf 1\{1\le k\le G-1\}
\sum_{i=1}^G
\left(
\frac{c_i}{k}-\frac{1}{G}
\right)
\nabla_\theta \log \pi_\theta(y_i|x).
\]

\begin{lemma}[Exact binary objective of Algorithm 1]
For binary rewards, the expected update of Algorithm 1 satisfies
\[
\mathbb E[\hat g_{\mathrm{RB}}(x)]
=
\nabla_\theta f_{G-1}(p_\theta(x)),
\]
where
\[
f_H(p)
=
\sum_{m=1}^{H}
\frac{1-(1-p)^m}{m},
\qquad H=G-1.
\]
Moreover, \(f_H\) is concave in \(p\).
\end{lemma}

\begin{proof}
Let
\[
\mu_+ := \mathbb E[\nabla_\theta \log \pi_\theta(y|x)\mid c(y)=1],
\qquad
\mu_- := \mathbb E[\nabla_\theta \log \pi_\theta(y|x)\mid c(y)=0].
\]
By the score-function identity,
\[
\mu_+ = \nabla_\theta \log p_\theta(x),
\qquad
\mu_- = \nabla_\theta \log(1-p_\theta(x)).
\]

Conditioned on \(k\in\{1,\ldots,G-1\}\), there are \(k\) correct samples and
\(G-k\) incorrect samples. Therefore,
\[
\mathbb E[\hat g_{\mathrm{RB}}(x)\mid k]
=
k\left(\frac{1}{k}-\frac{1}{G}\right)\mu_+
+
(G-k)\left(-\frac{1}{G}\right)\mu_-.
\]
This simplifies to
\[
\mathbb E[\hat g_{\mathrm{RB}}(x)\mid k]
=
\frac{G-k}{G}(\mu_+-\mu_-).
\]
Since \(k\sim\mathrm{Binomial}(G,p)\), we have
\[
\mathbb E\left[
\frac{G-k}{G}\mathbf 1\{1\le k\le G-1\}
\right]
=
(1-p)-(1-p)^G.
\]
Hence
\[
\mathbb E[\hat g_{\mathrm{RB}}(x)]
=
\big((1-p)-(1-p)^G\big)(\mu_+-\mu_-).
\]
Using
\[
\mu_+-\mu_-
=
\nabla_\theta \log p-\nabla_\theta \log(1-p)
=
\frac{\nabla_\theta p}{p(1-p)},
\]
we obtain
\[
\mathbb E[\hat g_{\mathrm{RB}}(x)]
=
\frac{1-(1-p)^{G-1}}{p}\nabla_\theta p.
\]
Therefore this is the gradient of a scalar objective \(f_{G-1}(p)\) satisfying
\[
f'_{G-1}(p)
=
\frac{1-(1-p)^{G-1}}{p}
=
\sum_{m=0}^{G-2}(1-p)^m.
\]
Integrating gives
\[
f_{G-1}(p)
=
\sum_{m=1}^{G-1}
\frac{1-(1-p)^m}{m}
\]
up to an additive constant.

Finally,
\[
f''_{G-1}(p)
=
-\sum_{m=1}^{G-2} m(1-p)^{m-1}\le 0,
\]
so \(f_{G-1}\) is concave.
\end{proof}
For fixed $p\in(0,1]$, let $H_n=\sum_{m=1}^n1/m$. Then
\[
f_n(p)-H_n=-\sum_{m=1}^n\frac{(1-p)^m}{m}\longrightarrow\log p.
\]
Thus the logarithmic limit is understood up to a parameter-independent additive constant; $f_n(p)$ itself does not have a finite limit. The score-function derivation assumes $0<p_\theta(x)<1$ and the usual differentiation-under-expectation conditions.
\subsection{Idealized Non-shifted Reward-normalized Estimator}
\label{app:oo_general}

\begin{assumption}[General reward setting]
\label{ass:general-reward}
Fix a prompt $x$. Let $y_1,\dots,y_G \stackrel{\text{i.i.d.}}{\sim}\pi_\theta(\cdot\mid x)$. Define
\[
R_i := R(y_i),\quad S := \sum_{j=1}^G R_j,\quad 
\mu_\theta(x):=\mathbb E_{y\sim\pi_\theta(\cdot\mid x)}[R(y)].
\]
\[\phi_i := \nabla_\theta\log\pi_\theta(y_i\mid x)\in\mathbb R^d.\]
Assume:
\begin{enumerate}
    \item There are fixed constants $0<r_{\mathrm{lo}}\le r_{\mathrm{hi}}<\infty$ such that $r_{\mathrm{lo}}\le R(y)\le r_{\mathrm{hi}}$ almost surely. These are sufficient conditions for this idealized estimator only; zero-valued binary rewards are covered separately by the exact binary analysis.
    \item $\mathbb E[\|\phi(y)\|^2\mid x]<\infty$, where $\phi(y)=\nabla_\theta\log\pi_\theta(y\mid x)$.
    \item Rewards have no explicit dependence on $\theta$, the policy has parameter-independent support locally, and differentiation may pass under the relevant expectations. Thus $\mathbb E[\phi]=0$ and $\mathbb E[R\phi]=\nabla_\theta\mu_\theta(x)$.
\end{enumerate}
All expectations below are conditional on $x$ and we omit the conditioning for brevity.
\end{assumption}

\vspace{0.2em}
\textbf{\textit{Asymptotic-G Derivation}}
\vspace{0.5em}

Define per-sample normalized rewards
\[
\tilde r_i=
\begin{cases}
R_i/S, & S>0,\\
0, & S=0,
\end{cases}
\]
and the VPG/REINFORCE~\citep{williams1992simple, NIPS1999_464d828b} group estimator
\[
\begin{aligned}
g_{\mathrm{new}}(x)
&:= \sum_{i=1}^G \tilde r_i\,\nabla_\theta\log\pi_\theta(y_i\mid x) \\
&= \mathbf 1[S>0]\sum_{i=1}^G \frac{R_i}{S}\,
   \nabla_\theta\log\pi_\theta(y_i\mid x).
\end{aligned}
\]

\begin{lemma}[large $G$ objective]
\label{lem:large-G}
Under the assumptions above,
\[
\lim_{G\to\infty}\mathbb E[g_{\mathrm{new}}(x)]
=
\nabla_\theta\log \mu_\theta(x).
\]
Consequently, for large $G$ the implicit objective is
\[
J_{\mathrm{new}} \approx \mathbb E_x[\log \mu_\theta(x)].
\]
\end{lemma}

\begin{proof}
Write $\bar R_G=G^{-1}\sum_iR_i$, $\bar\psi_G=G^{-1}\sum_iR_i\phi_i$,
$\mu=\mathbb E[R]$, and $A=\mathbb E[R\phi]=\nabla_\theta\mu$.
The estimator is $g_{\mathrm{new}}=\bar\psi_G/\bar R_G$ and
$\bar R_G\ge r_{\mathrm{lo}}>0$. Therefore
\[
g_{\mathrm{new}}-\frac{A}{\mu}
=\frac{(\bar\psi_G-A)-(A/\mu)(\bar R_G-\mu)}{\bar R_G}.
\]
Independence and the finite second moments give
\[
\mathbb E\left\|g_{\mathrm{new}}-\frac{A}{\mu}\right\|^2
\le \frac{2}{r_{\mathrm{lo}}^2G}
\left(\mathbb E\|R\phi-A\|^2+
\frac{\|A\|^2}{\mu^2}\operatorname{Var}(R)\right).
\]
Hence $g_{\mathrm{new}}\to A/\mu$ in $L^2$ and therefore in $L^1$.
Taking expectations yields $\mathbb E[g_{\mathrm{new}}]\to
\nabla_\theta\log\mu$. The lower reward bound controls the random denominator.
\end{proof}

\vspace{0.2em}
\textbf{\textit{Finite-G Approximations}}
\vspace{0.5em}

\begin{lemma}[$L^2$-rate of $g_G$ around $\nabla\log\mu$]
\label{lem:L2-rate}
Under Assumption~\ref{ass:general-reward}, there exists a finite constant $C(x)>0$ (depending on the distribution at prompt $x$ but not on $G$) such that for all sufficiently large $G$,
\[
\mathbb E\big[\|g_G(x) - \nabla_\theta\log\mu_\theta(x)\|^2\big]
\le \frac{C(x)}{G}.
\]
\end{lemma}

\begin{proof}
Write $g_G=g_{\mathrm{new}}$ for a group of size $G$. The direct ratio bound
in the proof of Lemma~\ref{lem:large-G} applies for every $G\ge1$.
The claim follows by choosing
\[
C(x)=\frac{2}{r_{\mathrm{lo}}^2}
\left(\mathbb E\|R\phi-A\|^2+
\frac{\|A\|^2}{\mu^2}\operatorname{Var}(R)\right),
\]
or any larger positive constant. It is finite under
Assumption~\ref{ass:general-reward}.
\end{proof}

\begin{corollary}[Bias, variance and MSE of $g_G$]
\label{cor:bias-var}
Under Assumption~\ref{ass:general-reward}, there exists $C(x)<\infty$ such that for all sufficiently large $G$,
\begin{align*}
\big\|\mathbb E[g_G(x)] - \nabla_\theta\log\mu_\theta(x)\big\|
&\le \sqrt{\frac{C(x)}{G}}
\end{align*}
\[\mathrm{Var}\big(g_G(x)\big)
:= \mathbb E\big[\|g_G(x) - \mathbb E[g_G(x)]\|^2\big]
\;\le\; \frac{C(x)}{G}\]
\end{corollary}

\begin{proof}
The mean-squared error bound is Lemma~\ref{lem:L2-rate}.
The bias bound follows from Cauchy--Schwarz:
\[
\big\|\mathbb E[g_G(x)] - \nabla_\theta\log\mu_\theta(x)\big\|^2
\le \mathbb E\big[\|g_G(x) - \nabla_\theta\log\mu_\theta(x)\|^2\big].
\]

For the variance,
\[
\mathrm{Var}\big(g_G(x)\big)
= \mathbb E\big[\|g_G(x) - \mathbb E[g_G(x)]\|^2\big]
\]
\[\le \mathbb E\big[\|g_G(x) - \nabla_\theta\log\mu_\theta(x)\|^2\big],\]
since $\nabla_\theta\log\mu_\theta(x)$ is a fixed vector.
The last quantity is bounded by $C(x)/G$ by Lemma~\ref{lem:L2-rate}.
\end{proof}

\subsection{Gap-normalized Estimator under General Rewards}
\label{app:oo_gap}

We now give sufficient conditions for the actual gap-normalized estimator.
Fix a prompt $x$ and policy $\theta$, and draw an i.i.d. sequence of rollouts.
Write $R_i=R(y_i,x)$, $\phi_i=\nabla_\theta\log\pi_\theta(y_i\mid x)$,
$M_G=\min_{i\le G}R_i$, and $Z_G=\sum_i(R_i-M_G)$.
Let $a(x)$ be the finite essential infimum of $R$ under $\pi_\theta$, assumed policy-independent in a neighborhood of $\theta$. Assume
$\mu=\mathbb E[R]>a(x)$, and
\[
\mathbb E|R|<\infty,\qquad
\mathbb E\|\phi\|<\infty,\qquad
\mathbb E[|R|\|\phi\|]<\infty.
\]
Assume also that rewards have no explicit $\theta$ dependence, the policy has
locally parameter-independent support, and differentiation may pass under the
expectation, so that $\mathbb E[\phi]=0$ and
$\mathbb E[R\phi]=\nabla_\theta\mu$.

For $Z_G>0$, the centered Algorithm~\ref{alg:garo} score-function estimator is
\[
\hat g_{\mathrm{gap}}
=\sum_{i=1}^G\left(\frac{R_i-M_G}{Z_G}-\frac1G\right)\phi_i;
\]
it is zero when $Z_G=0$. The centered term cannot in general be discarded
by treating the group-dependent skipping rule as an action-independent event.
Instead, with $\bar\phi_G=G^{-1}\sum_i\phi_i$, rewrite the estimator as
\[
\hat g_{\mathrm{gap}}
=\frac{G^{-1}\sum_iR_i\phi_i-M_G\bar\phi_G}
{G^{-1}\sum_iR_i-M_G}-\bar\phi_G
\quad\text{when }Z_G>0.
\]
The sampled minimum converges almost surely to $a(x)$. By the strong law,
the numerator converges to $\nabla_\theta\mu$, the denominator converges to
$\mu-a(x)>0$, and $\bar\phi_G\to0$. Thus
\[
\hat g_{\mathrm{gap}}\xrightarrow{a.s.}
\frac{\nabla_\theta\mu}{\mu-a(x)}
=\nabla_\theta\log(\mu-a(x)).
\]
Convergence of the expected update additionally requires uniform
integrability. One sufficient condition is an almost-sure bound
$\|\phi\|\le L<\infty$: the nonnegative normalized gaps sum to one, so
$\|\hat g_{\mathrm{gap}}\|\le2L$, and dominated convergence then applies.
Under these sufficient conditions, the large-group expected direction has a
shifted log-mean interpretation. This argument establishes neither an exact
finite-$G$ general-reward objective nor a finite-$G$ rate; the behavior of the
sampled minimum can depend on the reward distribution near its lower endpoint.
When $a(x)=0$, the limiting expression coincides with the unshifted log-mean.

\subsection{RL as Policy-to-Target KL Minimization}
\label{app:rl_forward}

\subsubsection{Trajectory Level}

\textit{Claim.} 
The regularized RL objective
\[
\begin{aligned}
J_{\mathrm{RL}}(\theta)
&= \mathbb{E}_{\gamma \sim \pi_\theta}\!\big[R(\gamma)\big]
- \beta\, D_{\mathrm{KL}}\!\big(\pi_\theta \,\|\, \pi_{\mathrm{ref}}\big)
\end{aligned}
\]
is equivalent (up to an additive constant independent of $\theta$) to
\[
\begin{aligned}
-\beta\,D_{\mathrm{KL}}\!\big(\pi_\theta \,\|\, \pi^*_{\mathrm{RL}}\big),
\end{aligned}
\]
where
\[
\begin{aligned}
\pi^*_{\mathrm{RL}}(\gamma)
&\propto
\pi_{\mathrm{ref}}(\gamma)\exp\!\left(\frac{R(\gamma)}{\beta}\right).
\end{aligned}
\]
\begin{proof}
Let $\gamma$ denote a trajectory (rollout) and let $\pi_\theta(\gamma)$ be the trajectory
distribution induced by policy $\pi_\theta$. Consider
\[
\begin{aligned}
J_{\mathrm{RL}}(\theta)
&= \mathbb{E}_{\gamma\sim\pi_\theta}[R(\gamma)]
-\beta\,D_{\mathrm{KL}}(\pi_\theta\|\pi_{\mathrm{ref}}).
\end{aligned}
\]
Expanding the KL term,
\[
\begin{aligned}
D_{\mathrm{KL}}(\pi_\theta\|\pi_{\mathrm{ref}})
&=\mathbb{E}_{\gamma\sim\pi_\theta}\!\left[
\log\frac{\pi_\theta(\gamma)}{\pi_{\mathrm{ref}}(\gamma)}
\right],
\end{aligned}
\]
so
\[
\begin{aligned}
J_{\mathrm{RL}}(\theta)
&=\mathbb{E}_{\gamma\sim\pi_\theta}\!\left[
R(\gamma)-\beta\log\pi_\theta(\gamma)
+\beta\log\pi_{\mathrm{ref}}(\gamma)
\right].
\end{aligned}
\]

Define the (nonparametric) optimal distribution
\[
\begin{aligned}
\pi^*_{\mathrm{RL}}(\gamma)
&=\frac{1}{Z}\,\pi_{\mathrm{ref}}(\gamma)
\exp\!\left(\frac{R(\gamma)}{\beta}\right),\\
Z
&=\sum_\gamma \pi_{\mathrm{ref}}(\gamma)
\exp\!\left(\frac{R(\gamma)}{\beta}\right).
\end{aligned}
\]
Then
\[
\begin{aligned}
\log \pi^*_{\mathrm{RL}}(\gamma)
&=\log\pi_{\mathrm{ref}}(\gamma)
+\frac{R(\gamma)}{\beta}-\log Z.
\end{aligned}
\]
Compute the policy-to-target KL:
\[D_{\mathrm{KL}}(\pi_\theta\|\pi^*_{\mathrm{RL}})\]
\[=\mathbb{E}_{\gamma\sim\pi_\theta}\!\left[
\log\pi_\theta(\gamma)-\log\pi^*_{\mathrm{RL}}(\gamma)
\right]\]
\[=\mathbb{E}_{\gamma\sim\pi_\theta}\!\left[
\log\pi_\theta(\gamma)-\log\pi_{\mathrm{ref}}(\gamma)
-\frac{R(\gamma)}{\beta}
+\log Z
\right]\]

Multiplying by $-\beta$ gives
\[-\beta D_{\mathrm{KL}}(\pi_\theta\|\pi^*_{\mathrm{RL}})=\mathbb{E}\!\left[
R(\gamma)-\beta\log\pi_\theta(\gamma)
+\beta\log\pi_{\mathrm{ref}}(\gamma)
\right]\]
\[-\beta\log Z
=J_{\mathrm{RL}}(\theta)-\beta\log Z.
\]
Therefore,
\[
\begin{aligned}
J_{\mathrm{RL}}(\theta)
&= -\beta D_{\mathrm{KL}}(\pi_\theta\|\pi^*_{\mathrm{RL}})
+\beta\log Z,
\end{aligned}
\]
and since $\log Z$ does not depend on $\theta$,
\[
\begin{aligned}
\arg\max_\theta J_{\mathrm{RL}}(\theta)
&\equiv
\arg\min_\theta D_{\mathrm{KL}}(\pi_\theta\|\pi^*_{\mathrm{RL}}).
\end{aligned}
\]
\end{proof}

\subsubsection{Action Level}

\textit{Claim.} 
For any state $s$, maximizing the per-state objective
\[
\begin{aligned}
\mathcal L_s(\pi)
:= \mathbb E_{a\sim\pi(\cdot|s)}
\Big[
f(s,a)
-\beta\log\frac{\pi(a|s)}{\pi_{\rm ref}(a|s)}
\Big]
\end{aligned}
\]
is equivalent to minimizing the policy-to-target KL
\[D_{\rm KL}(\pi(\cdot|s)\|\pi_f^*(\cdot|s)),\]
where
$\pi_f^*(a|s)\propto\pi_{\rm ref}(a|s)\exp(f(s,a)/\beta)$; moreover, replacing
$f$ by $f-b(s)$ leaves $\pi_f^*$ unchanged, while using $f=r(s,a)$ generally
does not recover the $f=Q(s,a)$ optimum unless $Q(s,a)=r(s,a)+c(s)$.

\begin{proof}
Fix a state $s$ and consider the per-state regularized objective
\[
\begin{aligned}
\mathcal L_s(\pi)
:= \mathbb E_{a\sim\pi(\cdot|s)}
\Big[
f(s,a)
-\beta\log\frac{\pi(a|s)}{\pi_{\rm ref}(a|s)}
\Big],
\end{aligned}
\]
where $f$ is any score function. Define
\[
\begin{aligned}
\pi_f^*(a|s)
&:= \frac{1}{Z_f(s)}\,\pi_{\rm ref}(a|s)\exp\!\Big(\frac{f(s,a)}{\beta}\Big),
\end{aligned}
\]
\[Z_f(s)=\sum_a \pi_{\rm ref}(a|s)e^{f(s,a)/\beta}.\]
Then
\[
\log \pi_f^*(a|s)=\log\pi_{\rm ref}(a|s)+\frac{f(s,a)}{\beta}-\log Z_f(s),
\]
and hence
\[D_{\rm KL}\!\big(\pi(\cdot|s)\,\|\,\pi_f^*(\cdot|s)\big)\]
\[=\mathbb E_{a\sim\pi}\!\left[
\log\pi(a|s)-\log\pi_f^*(a|s)
\right]\]
\[=\mathbb E_{a\sim\pi}\!\left[
\log\frac{\pi(a|s)}{\pi_{\rm ref}(a|s)}
-\frac{f(s,a)}{\beta}
+\log Z_f(s)
\right].\]

Multiplying by $-\beta$ gives
\[
\begin{aligned}
-\beta D_{\rm KL}(\pi\|\pi_f^*)
&=\mathcal L_s(\pi)-\beta\log Z_f(s).
\end{aligned}
\]
Since $\log Z_f(s)$ is independent of $\pi$, maximizing $\mathcal L_s(\pi)$ is
equivalent to minimizing this policy-to-target KL, and the unique maximizer is
$\pi_f^*$.

\textbf{Baseline invariance.}
Let $f'(s,a)=f(s,a)-b(s)$ for any state-only $b(s)$. Then
\[
\begin{aligned}
\pi_{f'}^*(a|s)
&\propto \pi_{\rm ref}(a|s)\exp\!\Big(\frac{f(s,a)-b(s)}{\beta}\Big)\\
&=\exp\!\Big(-\frac{b(s)}{\beta}\Big)
\pi_{\rm ref}(a|s)e^{f(s,a)/\beta}
\propto \pi_f^*(a|s),
\end{aligned}
\]
so $\pi_{f'}^*=\pi_f^*$. Taking $f=Q$ and $b=V$ yields
$\pi^*(a|s)\propto \pi_{\rm ref}(a|s)e^{A(s,a)/\beta}$, identical to using $Q$.

\textbf{Using $r(s,a)$.}
Replacing $f$ by the immediate reward gives
$\pi_r^*(a|s)\propto \pi_{\rm ref}(a|s)e^{r(s,a)/\beta}$, which coincides with
$\pi_Q^*$ iff $Q(s,a)=r(s,a)+c(s)$ for some state-only $c(s)$. This holds only in
special cases (e.g., contextual bandits or $\gamma=0$); in general MDPs
($\gamma>0$), $Q$ depends on future returns, so $\pi_r^*\neq \pi_Q^*$.
\end{proof}

\section{Experimental Detail}
\label{app:hyper}

\subsection{Data Processing}

\paragraph{Training \& Validation.} We filter DAPO-MATH-17K \citep{yu2025dapoopensourcellmreinforcement} using the following criteria:
\begin{enumerate}
    \item the final answer is an integer, so that we can reliably extract and compare answers from RL rollouts;
    \item the prompt length is at most 1024 tokens;
    \item the response length is at most 8192 tokens.
\end{enumerate}

After filtering, we then use the first 10k examples for training and the subsequent 1k examples for validation.

\paragraph{Evaluation.} For all benchmarks, we apply no filtering and keep the full dataset.

\subsection{Hyper-Parameters}

\paragraph{Training.} All RL algorithms were trained using the same hardware allocation. We sample 8 rollouts per prompt at temperature 1.0, with a maximum response length of 8192 tokens. We use AdamW with weight decay $0.1$ and a constant learning rate of $1\times 10^{-6}$ with no warmup, and perform one update per batch of 32 prompts. For the main benchmark comparisons, we retain each baseline\textquotesingle{}s KL setting when adding reward budgeting. GRPO uses a reference-policy KL coefficient of $0.001$ in both cases. 

\paragraph{Validation.} We validate every 5 training steps by computing pass@1 accuracy on the validation set. For generation, we use greedy decoding with a maximum response length of 8192 tokens to avoid stochasticity.

\paragraph{Evaluation.} We evaluate each trained checkpoint on the selected reasoning benchmarks using $8$ responses per prompt at temperature $0.7$, with a maximum response length of $8192$ tokens. As in Table~\ref{tab:main_results}, we report mean accuracy across the eight generations for Qwen3-8B and pass@8 for Llama3.2-3B-Instruct. These are different metrics and should not be compared directly across model families. The table reports point estimates rather than $\text{mean}\pm\text{std}$ across independently trained models. For the integer-answer benchmarks AMC23, AIME24, and AIME25, we use exact match. For MATH500 and OlympiadBench, whose answers may be symbolic expressions, we use an LLM judge (Qwen3-235B-Instruct) to determine whether a predicted answer matches the ground truth. The judge is instructed to compare \emph{mathematical meaning} rather than surface form: it should ignore differences in \LaTeX{}, whitespace, and notation, and accept mathematically equivalent expressions.
It must also enforce any required domain, units, and exactness constraints, while allowing small rounding errors when appropriate.
The judge outputs \emph{only} one of two labels: $\boxed{\text{correct}}$ or $\boxed{\text{incorrect}}$.

\noindent\textbf{System message (judge instruction).}
\begin{Verbatim}[fontsize=\footnotesize, breaklines=true, breakanywhere=true,
                 frame=single, framesep=2mm, xleftmargin=0.5em]
Judge if the Predicted Answer matches the Ground Truth for the Given Problem.

Be strict on meaning, not format: ignore LaTeX/whitespace/notation; accept mathematically equivalent forms.
Respect required domain/units/exactness; allow small rounding error.

Reply ONLY with \boxed{correct} or \boxed{incorrect}.
\end{Verbatim}

\noindent\textbf{User message (template).}
\begin{Verbatim}[fontsize=\footnotesize, breaklines=true, breakanywhere=true,
                 frame=single, framesep=2mm, xleftmargin=0.5em]
Given Problem: <question>
Ground Truth: <ground_truth>
Predicted Answer: <pred_answer>
\end{Verbatim}

\subsection{Convergence Step Detection}
\label{app:convergence_step_detection}

In Figure~\ref{fig:sharpness_factor}, we retrospectively define an operational “convergence step” for each configuration based on a smoothed trajectory rather than raw, potentially noisy evaluations. Concretely, given the sequence of validation accuracies measured every 5 steps, we fit a uni-variate polynomial ($p(x)$) (degree $d=5$) to $(x,y)$ with $x$ the training step and $y$ the scaled validation accuracy, and evaluate the fitted curve on a dense grid $\{x_g\}_{g=1}^{N_{\mathrm{grid}}}$ ($N_{\mathrm{grid}}=400$) to obtain ${y_g=p(x_g)}$. We then identify the earliest grid point $x_g$ that satisfies a sustained plateau condition: (i) local flatness, meaning the absolute change over a forward window of length $w=0.10N_{\mathrm{grid}}$ (a fixed fraction of the grid) is small, $|y_{g+w}-y_g|\le \varepsilon_{\text{local}}$; and (ii) global closeness, meaning the remaining gain to the future best value on the fitted curve is small, $\max_{j\ge g} (y_j - y_g) \le \varepsilon_{\text{remain}}$. The tolerances are set relative to the fitted curve’s dynamic range $\Delta=\max_g y_g-\min_g y_g$: $\varepsilon_{\text{local}}=\max(\varepsilon_{\text{abs}}, \varepsilon_{\text{rel}}\Delta)$ with $\varepsilon_{\text{abs}}=0.005$, $\varepsilon_{\text{rel}}=0.02$, and $\varepsilon_{\text{remain}}=\max(\varepsilon_{\text{abs}}, \rho\Delta)$ with $\rho=0.10$. To avoid spurious detections, we additionally require that both criteria hold continuously for at least $s$ grid points with $s=0.06N_{\mathrm{grid}}$ and a minimum of 10 points. The convergence step is reported as the earliest $x_g$ that meets this sustained condition; we visualize it as a vertical dashed line at $x=x_g$, and optionally plot a plateau guide as the median of the last 10\% of fitted values. This procedure measures plateau timing retrospectively on the plotted trajectories. Training uses the validation-plateau stopping rule described above; the numerical tolerances here specify only this retrospective diagnostic.

\section{Conditional Association Analysis}
\label{app:deconfounding_regression}

Under the set-up in \S\ref{sec:sharpness_factor_controlled_studies}, we further test whether entropy’s predictiveness survives after controlling for $\kappa$. Specifically, for every run with the principled sharpness factor $\kappa$ and update step $t$, we record the entropy $H_{\kappa,t}$ and the validation accuracy $P_{\kappa,t}$. We then set a prediction horizon $\Delta=10\text{ steps}$ and regress twice to test whether $H_{\kappa,t}$ predicts $P_{\kappa,t+\Delta}$.

\begin{figure}[htbp]
    \centering
    \includegraphics[width=0.7\linewidth]{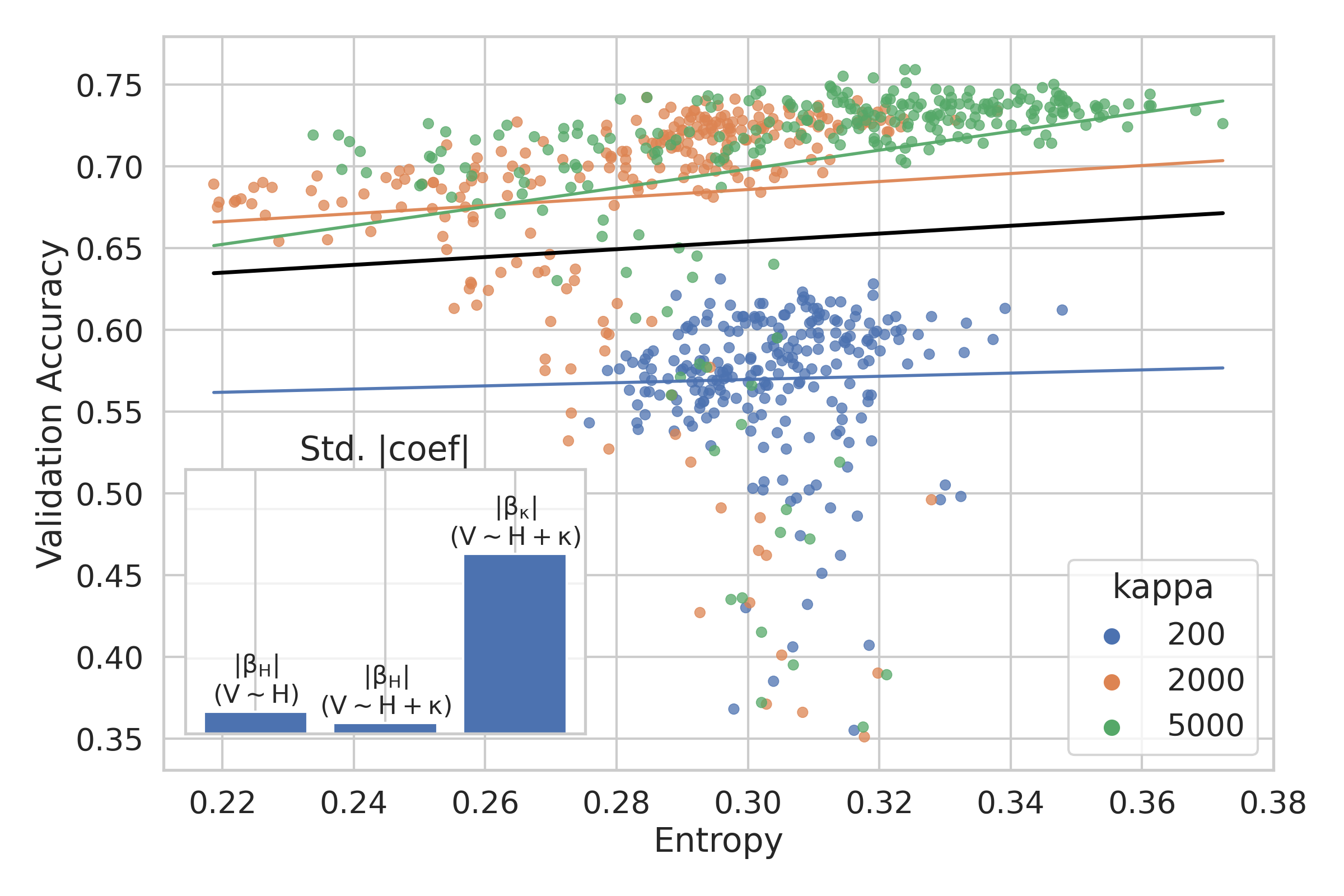}
    \caption{Accuracy evaluated at predictive horizon $10$ is plotted against the entropy for three settings $\kappa=200,2000,5000$. Different marker
    groups indicate different $\kappa$. The inset describes the
    absolute regression coefficients $|\beta_H|$ from OLS linear fits that relate the
    target metric ``Validation Accuracy'' $V$ to (i) ``Entropy'' $H$ alone and
    (ii) $H$ together with $\kappa$, showing the relative importance of $H$ vs. $\kappa$.}
    \label{fig:entropy_mediator}
\end{figure}

First, we fit a naive regression
$
P_{\kappa,t+\Delta} = a + \beta_H\, H_{\kappa,t} + \varepsilon_{\kappa,t}.
$
Next, we include the configuration-level quantity $\kappa$ via
$
P_{\kappa,t+\Delta} = a + \beta_H'\, H_{\kappa,t} + \beta_\kappa\, \kappa + \varepsilon_{\kappa,t}.
$
As shown in Figure~\ref{fig:entropy_mediator}, the points cluster much more clearly by $\kappa$ than by $H$. Consistently, once $\kappa$ is included as a predictor, it explains most of the predictive power, while the entropy coefficient shrinks substantially ($|\beta_H'|\ll|\beta_H|$). Thus, in this controlled comparison, much of the apparent predictive signal of entropy is accounted for by variation in $\kappa$. This comparison describes conditional association in the pooled trajectories. It does not control for training-step effects or serial dependence, and the configurations vary reward scale and KL strength jointly; it therefore does not identify a causal effect of $\kappa$ or establish entropy as a downstream mediator.

\section{Limitations}
\label{sec:limitations}
Our experiments cover verifiable reasoning tasks with predominantly binary rewards; the general-reward analysis assumes additional regularity and does not establish performance for noisy, learned, or open-ended rewards. Each main configuration uses one training seed and $G=8$, with no group-size sweep. Scores are point estimates; eight evaluation responses measure decoding variability, not training-seed uncertainty. The finite-horizon interventions and regressions do not establish a unique causal mediator of eventual performance. 
% Training stops at a validation-accuracy plateau; numerical thresholds, per-run training lengths, and main-table checkpoint-selection rules are unreported. 
Appendix~\ref{app:convergence_step_detection} retrospectively measures convergence in the Figure~\ref{fig:sharpness_factor} intervention trajectories.

\end{document}

%% file: math_commands.tex
\usepackage{amsmath,amsfonts,bm}
\usepackage{natbib}
\usepackage{fancyhdr}
\def\eqref#1{equation~\ref{#1}}
\def\1{\bm{1}}

\DeclareMathAlphabet{\mathsfit}{\encodingdefault}{\sfdefault}{m}{sl}
\SetMathAlphabet{\mathsfit}{bold}{\encodingdefault}{\sfdefault}{bx}{n}